\documentclass[runningheads]{llncs}
\usepackage{graphicx, algorithm, float}
\usepackage{amsmath,amssymb,mathtools} 
\usepackage{enumitem}
\usepackage{color}
\usepackage{subfig}
\usepackage[noend]{algpseudocode}
\usepackage{fixltx2e}
\usepackage[width=122mm,left=12mm,paperwidth=146mm,height=193mm,top=12mm,paperheight=217mm]{geometry}

\def\NoNumber#1{{\def\alglinenumber##1{}\State #1}\addtocounter{ALG@line}{-1}}

\DeclareMathOperator*{\argmin}{\arg\!\min}
\newcommand{\etal}{\textit{et al}.}
\newcommand{\ie}{\textit{i}.\textit{e}.}

\DeclarePairedDelimiter\norm{\lVert}{\rVert}

\begin{document}
\pagestyle{headings}
\mainmatter

\title{Joint Projection and Dictionary Learning using Low-rank Regularization and Graph Constraints} 

\titlerunning{Joint Projection and Dictionary Learning ...}

\authorrunning{Homa Foroughi, Nilanjan Ray, Hong Zhang}

\author{Homa Foroughi, Nilanjan Ray, Hong Zhang}
\institute{University of Alberta}

\maketitle

\begin{abstract}
In this paper, we aim at learning simultaneously a discriminative dictionary and a robust projection matrix from noisy data. The joint learning, makes the learned projection and dictionary a better fit for each other, so a more accurate classification can be obtained. However, current prevailing joint dimensionality reduction and dictionary learning methods, would fail when the training samples are noisy or heavily corrupted. 
To address this issue, we propose a joint projection and dictionary learning using low-rank regularization and graph constraints (JPDL-LR). 
Specifically, the discrimination of the dictionary is achieved by imposing Fisher criterion on the coding coefficients. In addition, our method explicitly encodes the local structure of data by incorporating a graph regularization term, that further improves the discriminative ability of the projection matrix. Inspired by recent advances of low-rank representation for removing outliers and noise, we enforce a low-rank constraint on sub-dictionaries of all classes to make them more compact and robust to noise. 
Experimental results on several benchmark datasets verify the effectiveness and robustness of our method for both dimensionality reduction and image classification, especially when the data contains considerable noise or variations.
\end{abstract}
\setlength{\belowdisplayskip}{4pt} \setlength{\belowdisplayshortskip}{0pt}
\setlength{\abovedisplayskip}{4pt} \setlength{\abovedisplayshortskip}{0pt}
\raggedbottom
\section{Introduction}
\label{sec:intro}
Sparse representation of signals has attracted tremendous interest and has been successfully applied to many computer vision applications \cite{SRC}. According to sparse representation theory, signals can be well-approximated by linear combination of a few columns of some appropriate basis or dictionary. The dictionary, which should faithfully and discriminatively represent the encoded signal, plays an important role in the success of sparse representation \cite{FDDL} and it has been shown that learned dictionaries significantly outperform pre-defined ones such as Wavelets \cite{Wavelet}. The last few years have witnessed fast development on dictionary learning (DL) approaches and great success has been demonstrated in different computer vision applications such as image classification.

Moreover, in many areas of computer vision and pattern recognition, data are characterized by high dimensional feature vectors; however, dealing with high-dimensional data is challenging for many tasks such as DL. High-dimensional data are not only inefficient and computationally intensive, but the sheer number of dimensions often masks the discriminative signal embedded in the data \cite{LGE-KSVD}. Therefore, finding a low-dimensional projection seems to be a natural solution. In general, dimensionality reduction (DR) techniques map data to a low-dimensional space such that non-informative and irrelevant information of data are discarded \cite{SE}. Usually, DR is performed first to the training samples and the dimensionality reduced data are used for DL. However, recent studies reveal that the pre-learned DR matrix neither fully promotes the underlying structure of data \cite{SE}, nor preserves the best features for DL \cite{JDDRDL}. Intuitively, the DR and DL processes should be jointly conducted for a more effective classification.

Only a few works have discussed the idea of jointly learning the transformation of training samples and dictionary. Some of these techniques such as \cite{DR-SRC}, \cite{sparse-linear} assume that the dictionary is given and cannot help the process of learning the dictionary. By addressing this issue, \cite{Simul-SRC} presented a simultaneous projection and DL method using a carefully designed sigmoid reconstruction error. The data is projected to an orthogonal space where the intra- and inter-class reconstruction errors are minimized and maximized, respectively for making the projected space discriminative. However, \cite{Invariance} showed that the dictionary learned in the projected space is not more discriminative than the one learned in the original space. 
JDDLDR method \cite{JDDRDL} jointly learns a DR matrix and a discriminative dictionary and achieves promising results for face recognition. The discrimination is enforced by a Fisher-like constraint on the coding coefficients, but the projection matrix is learned without any discrimination constraints. Nguyen \etal \, \cite{SE} proposed a joint DR and sparse learning framework by emphasizing on preserving the sparse structure of data. Their method, known as sparse embedding (SE) can be extended to a non-linear version via kernel tricks and also adopts a novel classification schema leading to great performance. Nevertheless, it fails to consider the discrimination power among the separately learned class-specific dictionaries, such that it is not guaranteed to produce improved classification performance \cite{Problem-SE}. Ptucha \etal \, \cite{LGE-KSVD} integrated manifold-based DR and sparse representation within a single framework and presented a variant of the K-SVD algorithm by exploiting a linear extension of graph embedding (LGE). The LGE concept is further leveraged to modify the K-SVD algorithm for co-optimizing a small, yet over-complete dictionary, the projection matrix and the coefficients. Most recently, Liu \etal \, \cite{JNPDL} proposed a joint non-negative projection and DL method. The discrimination is achieved by imposing graph constraints on both projection and coding coefficients that maximises the intra-class compactness and inter-class separability. 

Although, some of the aforementioned methods perform well for different classification and recognition tasks, the performance of these methods deteriorates when the training data are contaminated heavily because of occlusion, disguise, lighting variations or pixel corruption. In the recent years, low-rank (LR) matrix recovery, which efficiently removes noise from corrupted observations, has been successfully applied to a variety of computer vision applications, such as subspace clustering \cite{LR-Subspace}, background subtraction \cite{BS-LR} and image classification \cite{Homa-BMVC}. Accordingly, some DL methods have been proposed by integrating rank minimization into sparse representation that have achieved impressive results, especially when corruption exists \cite{D2L2R2}, \cite{Structured-LR-DL}. 

In this paper, we propose a novel framework, called \textit{joint projection and dictionary learning using low-rank regularization and graph constraints} (JPDL-LR), which brings the strength of both DL and LR together for an efficient DR. The algorithm learns a discriminative structured dictionary in the reduced space, whose atoms have correspondence to the class labels and Fisher discrimination criterion is imposed on the coding vectors to enhance class discrimination. Simultaneously, we consider optimizing the input feature space by jointly learning a feature projection matrix. In particular, a supervised nearest neighbor graph is built to encode the local structure information of data; consequently, the desirable relationship among training samples is preserved. To learn effective features from noisy data, we incorporate LR regularization into JPDL-LR objective function and impose a LR constraint on sub-dictionaries to make them robust to noise. This joint framework empowers our algorithm with several important advantages: (1) Learning in the reduced dimensions with lower computational complexity, (2) Ability to handle noisy and corrupted observations, (3) Maintaining both global and local structure of data, and (4) Promoting the discriminative ability of the learned projection and dictionary that is highly desired when the ultimate goal is classification. Extensive experimental results validate the effectiveness of our method for DL and DR and its applicability to image classification task, especially for noisy observations. 

The remainder of the paper is organized as follows. Section \ref{sec:proposed} presents the proposed JPDL-LR method. The optimization algorithms are described in Section \ref{sec:opt} and the classification scheme is explained in Section \ref{sec:classification}. Section \ref{sec:experiment} shows experimental results on different datasets and we draw conclusions in Section \ref{sec:conclusion}.
\raggedbottom
\section{The Proposed JPDL-LR Framework}
\label{sec:proposed}
We aim to learn a discriminative dictionary and a projection matrix simultaneously, using LR regularization and graph constraints. In this paper, we adopt D\textsuperscript{2}L\textsuperscript{2}R\textsuperscript{2} \cite{D2L2R2} framework due to its discrimination power and promising performance on noisy data. Let $X$ be a set of $m$-dimensional training samples, \ie, $X=\{ X_1, X_2, \dots , X_K \}$, where $X_i$ denotes the training samples from class $i$ and $K$ is the number of classes. The learned, class-specific structured dictionary is denoted by $D=\{ D_1, D_2, \dots , D_K \}$, where $D_i$ is the sub-dictionary associated with class $i$. 
We also want to learn the projection matrix $P \in R^{m \times d} \, (d<m)$, that projects data into a low-dimensional space. Denote by $A$ the sparse representation matrix of $P^T X$ over $D$, \ie, $P^T X \approx DA$. We can write $A$ as $A=\{ A_1, A_2, \dots , A_K \}$, where $A_i$ is the representation of $P^T X_i$ over $D$. Therefore, we propose the following optimization model:
\begin{gather}
\begin{aligned}
\hspace{-18pt} J_{(P,D,A)} = \argmin_{P,D,A} \Big\{ R(P,D,A) + \lambda_{1} \, \norm[\big]{A}_{1} + \lambda_{2} \, F(A) +  \alpha \sum\limits_{i}  \norm[\big]{D_i}_{*}  +  \delta \, G(P) \Big\}  
\label{eq1}
\end{aligned}
\end{gather}
where $R(P,D,A)$ is the discriminative reconstruction error, $\norm{A}_{1}$ denotes the $l_1$-regularization on coding coefficients, $F(A)$ is the Fisher discriminant function of the coefficients, $\norm{D_i}_{*}$ is the nuclear norm of sub-dictionary $D_i \,$, $G(P)$ represents the graph-based projection term and $\lambda_{1}, \lambda_{2}, \alpha, \delta$ are the scalar parameters.
\begin{enumerate} [label=\textbf{(\Alph*)},leftmargin=0.5cm]
\setlength\itemsep{0.3em}
\item \textbf{Discriminative Reconstruction Error}: To learn a representative and discriminative structured dictionary, each sub-dictionary $D_i$ should be able to well represent the dimensionality reduced samples from the $i$th class, but not other classes. To illustrate this idea mathematically, we rewrite $A_i$ as $A_i = [A_i^1 ; \dots; A_i^j ; \dots ; A_i^K]$, where $A_i^j$ is the representation coefficients of $P^T X_i$ over $D_j$. Our assumption implies that $A_i^i$ should have significant coefficients such that $\norm{P^T X_i - D_i A_i^i}_{F}^2$ is small, while for samples from class $j \, (j \neq i)$, $A_i^j$ should have nearly zero coefficients, such that $\norm{D_j A_i^j}_{F}^2$ is as small as possible. Moreover, the whole dictionary $D$ should well represent dimensionality reduced samples from any class, which implies the minimization of $\norm{P^T X_i - D A_i}_{F}^2$ in our model. Thus, the discriminative reconstruction function is defined as:
\begin{gather}
\begin{aligned}
\hspace{-18pt} R(P,D,A) = \sum\limits_{i=1}^K \Big( \norm[\big]{P^T X_i - D A_i}_{F}^2 + \norm[\big]{P^T X_i - D_i A_i^i}_{F}^2 + \sum\limits_{j=1 , \atop j \neq i}^K \norm[\big]{D_j A_i^j}_{F}^2 \Big)
\label{eq2}
\end{aligned} 
\end{gather}
\item \textbf{Fisher Discriminant Function}: To further increase the discrimination capability of dictionary $D$, we enforce the coding coefficient matrix $A$ to be discriminative. This can be achieved by minimizing the within-class scatter of $A$, denoted by $S_W(A)$ and maximizing the between-class scatter of $A$, denoted by $S_B(A)$. These scatter matrices are defined as follows:
\begin{gather}
\begin{aligned}
\hspace*{-0.5 cm}S_W(A) = \sum\limits_{i=1}^K \sum\limits_{a_k \in A_i} (a_k - m_i) (a_k - m_i)^T, \,\,
S_B(A) = \sum\limits_{i=1}^K n_i (m_i - m) (m_i - m)^T \hspace{35pt} 
\label{eq3}
\end{aligned}
\end{gather}
where $m_i$ and $m$ are mean vectors of $A_i$ and $A$ respectively, and $n_i$ is the number of samples in the $i$th class. The Fisher criterion is defined as:
\begin{align}
F(A) = tr \big( S_W(A) \big) - tr \big( S_B(A) \big) + \eta  \norm[\big]{A}_{F}^2 
\label{eq4}
\end{align}
where $\eta$ is a scalar parameter and the regularization term $\norm{A}_{F}^2$ is introduced to make $F(A)$ smoother and convex \cite{FDDL}.
\item \textbf{Low-rank Regularization}: The training samples in each class are linearly correlated in many situations and reside in a low-dimensional subspace. So, the sub-dictionary $D_i$, which is representing data from the $i$th class, is reasonably low-rank. Imposing LR constraints on sub-dictionaries would make them compact and also mitigate the influence of noise \cite{D2L2R2}. To find the most compact bases, we need to minimize $\norm{D_i}_{*}$ for all classes in our optimization.
\item \textbf{Graph-based Projection Term}: We aim to learn a projection matrix that can preserve useful information and map the training samples to a discriminative space, where different classes are more discriminant toward each other. Using the training data matrix $X$ and its corresponding class label set, a fully connected supervised neighborhood graph of input space is constructed. Let $W$ be the weight matrix of the graph, if $x_i$ has the same class label as $x_j$ and meanwhile $x_i$ is amongst the $k$-nearest neighbors of $x_j$ (or vice versa), then $W_{ij} = exp (-\norm{x_i - x_j}^2/t)$; otherwise $W_{ij} =0$. The goal of graph embedding is to preserve the similarities amongst high-dimensional neighbors in the low-dimensional space. Note that supervised graph would enable us to preserve desirable relationship among training samples, even if they are corrupted and Euclidean distance cannot determine their neighborhood as an initial metric.

If $x_i$ and $x_j$ lie in the same subspace, their corresponding low-dimensional embeddings $y_i$ and $y_j$ should be near each other. This would preserve the local structure information of data and the discrimination information of different classes. Therefore, we define a cost function as $\frac{1}{2} \sum_{\substack{i,j=1}}^N W_{ij} \norm{y_i - y_j}_{2}^2$. Let $D$ be a diagonal matrix of column sums of $W$, $D_{ii} = \sum\nolimits_{j} W_{ij}$ and $L$ be the Laplacian matrix; then $L = D-W$. So, the cost function can be reduced to:
\begin{align}
P^* = \argmin_{P} \> tr(P^T X L X^T P) \quad s.t. \quad  P^T X D X^T P=I
\label{eq6}
\end{align}
We note that the constraint $P^T X D X^T P =I$ removes the arbitrary scaling factor in the embedding. In order to make the constraint simpler, here we use the definition of the normalized graph Laplacian \cite{Norm-Lap} as $\hat{L} = I - D^{-\frac {1}{2}} W D^{-\frac {1}{2}}$. Consequently, Eq.\ref{eq6} is reformulated as:
\begin{align}
P^* = \argmin_{P} \> tr(P^T X \hat{L} X^T P)  \quad  s.t. \quad  P^T P=I
\label{eq7}
\end{align}
\item \textbf{The JPDL-LR model}: By incorporating Eqs. \ref{eq2}, \ref{eq4} and \ref{eq7} into the main optimization model, we have the following JPDL-LR model:
\begin{gather}
\hspace*{-0.6 cm}J_{(P,D,A)} = \argmin_{P,D,A} \bigg\{ \sum\limits_{i=1}^K \Big( \norm[\big]{P^T X_i - D A_i}_{F}^2 + \norm[\big]{P^T X_i - D_i A_i^i}_{F}^2 + \sum\limits_{j=1 , \, j \neq i}^K \norm[\big]{D_j A_i^j}_{F}^2 \Big)    \,\,\,\,\,\,\,\,\,\,\,\,\,\,\,\,\,
\nonumber
\\
\hspace*{-0.7 cm}+ \lambda_{1} \, \norm[\big]{A}_{1} + \lambda_{2} \, \Big( tr \big( S_W(A)-S_B(A) \big) + \eta  \norm[\big]{A}_{F}^2 \Big) 
+ \alpha \sum\limits_{i=1}^K  \norm[\big]{D_i}_{*}  +  \delta \, tr(P^T X \hat{L} X^T P) \bigg\} \,\,\,\,\,\,\,\,\,\,
\nonumber
\\
 s.t. \quad P^T P=I \quad \text{and} \quad \norm[\big]{d_n}_{2} =1, \forall n \hspace{60pt}
\label{eq8}
\end{gather}
\end{enumerate}
\raggedbottom
\section{Optimization}
\label{sec:opt}
Although the objective function in Eq. \ref{eq8} is not jointly convex to $(P,D,A)$, it is convex with respect to each of $P$,$D$ and $A$ when the others are fixed. We adopt a standard iterative learning framework to jointly learn them in three major steps.
\subsection{Update of Coding Coefficients $A$}
Assuming that $D$ and $P$ are fixed, the objective function in Eq. \ref{eq8}, is reduced to sparse coding problem to compute $A=\{ A_1, A_2, \dots , A_K \}$. We update $A_i$ class-by-class and meanwhile, make all other $A_j (j \neq i)$ fixed. As a result, Eq. \ref{eq8} is further reduced to:
\begin{align}
J_{(A_i)} = \argmin_{A_i} \Big\{ R(P,D,A_i) +  \lambda_{1} \, \norm[\big]{A_i}_{1} + \lambda_{2} \, F_i(A_i) \Big\}
\label{eq9}
\end{align}
where $R(P,D,A_i) = \norm{P^T X_i - D A_i}_{F}^2 + \norm{P^T X_i - D_i A_i^i}_{F}^2 + \sum_{\substack{j=1, j \neq i}}^K \norm{D_j A_i^j}_{F}^2$ and $F_i(A_i) = \norm{A_i - M_i}_{F}^2 - \sum_{\substack{k=1}}^K \norm{M_k - M}_{F}^2 + \eta \norm{A_i}_{F}^2$. $M_k$ and $M$ are the mean vector matrices of class $k$ and all classes respectively \cite{FDDL}. Eq. \ref{eq9} can be solved using iterative projection method \cite{IPM} by rewriting it as:
\begin{align}
J_{(A_i)} = \argmin_{A_i} \big\{ Q(A_i) +  2 \tau \, \norm[\big]{A_i}_{1} \big\}
\label{eq10}
\end{align}
where $Q(A_i) = R(P,D,A_i) + \lambda_{2} \, F_i(A_i)$ and $\tau = \lambda_1/2$. More details are in \cite{IPM}.
\subsection{Update of Dictionary $D$}
Then, we optimize $D$, when $A$ and $P$ are fixed. We also update $D_i$ class-by-class, by fixing all other $D_j ( j \neq i)$. Similar to \cite{D2L2R2}, when $D_i$ is updated, the coding coefficients of $ P^T X_i$ over $D_i$, \ie, $A_i^i$ should also be updated to reflect this change. By ignoring irrelevant terms, the objective function of Eq.\ref{eq8} then reduces to:
\begin{gather}
\begin{aligned}
J_{(D_i)} = \argmin_{D_i,A_i^i} \Big\{ \norm[\big]{P^T X_i - D_i A_i^i - \sum\limits_{j=1 , \, j \neq i}^K D_j A_i^j}_{F}^2 + \norm[\big]{P^T X_i - D_i A_i^i}_{F}^2 
\\
+ \sum\limits_{j=1 , \, j \neq i}^K \norm[\big]{D_j A_i^j}_{F}^2 + \alpha \sum\limits_{i=1}^K  \norm[\big]{D_i}_{*} \Big\} \hspace{60pt}
\end{aligned}
\label{eq11}
\end{gather}
Denote $r(D_i) = \norm{P^T X_i - D_i A_i^i - \sum_{\substack{j=1, \, j \neq i}}^K D_j A_i^j}_{F}^2 \> + \sum_{\substack{j=1, j \neq i}}^K \norm{D_j A_i^j}_{F}^2 \,$, the objective function of Eq.\ref{eq11} is reformulated as:
\begin{gather}
\hspace*{-0.6 cm} \min_{D_i,A_i^i,E_i} \norm[\big]{A_i^i}_{1} + \alpha \norm[\big]{D_i}_{*} + \beta \, \norm[\big]{E_i}_{2,1} + \lambda \, r(D_i)
\quad s.t. \,\,\, P^T X_i = D_i A_i^i + E_i
\label{eq12}
\end{gather}
To facilitate the optimization, we introduce two relaxation variables $J$ and $Z$ and then Eq.\ref{eq12} can be rewritten as:
\begin{gather}
\begin{aligned}
\min_{D_i,A_i^i,E_i} \norm[\big]{Z}_{1} + \alpha \norm[\big]{J}_{*} + \beta \, \norm[\big]{E_i}_{2,1} + \lambda \, r(D_i)
\\
s.t. \quad P^T X_i = D_i A_i^i + E_i , \> D_i = J, \> A_i^i = Z 
\label{eq13}
\end{aligned}
\end{gather}
The above problem can be solved by inexact Augmented Lagrange Multiplier (ALM) method \cite{ALM}. The augmented Lagrangian function of Eq.\ref{eq13} is:
\begin{gather}
\begin{aligned}
\min_{D_i,A_i^i,E_i} \norm[\big]{Z}_{1} + \alpha \norm[\big]{J}_{*} + \beta \, \norm[\big]{E_i}_{2,1} + \lambda \, r(D_i) \hspace{65pt}
\\
+ tr \big[ T_1^T (P^T X_i - D_i A_i^i - E_i) \big] + tr \big[ T_2^T (D_i - J) \big] + tr \big[ T_3^T (A_i^i - Z) \big] \hspace{23pt}
\\
+ \frac{\mu}{2} \Big( \norm[\big]{P^T X_i - D_i A_i^i - E_i}_{F}^2 + \norm[\big]{D_i - J}_{F}^2 + \norm[\big]{A_i^i - Z}_{F}^2 \Big) \hspace{30pt}
\label{eq14}
\end{aligned}
\end{gather}
where $T_1$,$T_2$ and $T_3$ are Lagrange multipliers and $\mu$ is a balance parameter. The details of solving of Eq.\ref{eq14} can be found in Algorithm~\ref{alg1-update-di}.
\begin{algorithm}[h]
\scriptsize
\caption{Inexact ALM Algorithm for Eq.\ref{eq14}}
\label{alg1-update-di}
\begin{algorithmic}[1]
\renewcommand{\algorithmicrequire}{\textbf{Input:}}
\renewcommand{\algorithmicensure}{\textbf{Output:}}
\algnewcommand{\Initialize}{\State \textbf{Initialize:}}
\Require Reduced-dimensionality data $P^TX_i$, Sub-dictionary $D_i$, parameters $\alpha, \beta, \lambda$
\Ensure $D_i, E_i, A_i^i$
\Initialize $\, J=0, \, E_i=0, \, T_1=0, \, T_2=0, \, T_3=0, \, \mu=10^{-6}, \, max_{\mu} = 10^{30}, \, \epsilon = 10^{-8}, \, \rho =1.1 $
\While {not converged}
    \State Update $Z$ as: $\hspace{6pt} Z = \argmin_{Z} \, \Big( \frac{1}{\mu} \norm[\big]{Z}_{1} + \frac{1}{2} \norm[\big]{Z - (A_i^i + \frac{T3}{\mu})}_{F}^2 \Big) $
    \vspace{1mm}    
    \State Update $A_i^i$ as: $A_i^i = \Big( D_i^TD_i +I \Big)^{-1} \Big( D_i^T(P^T X_i - E_i) +Z + \frac{D_i^T T_1 - T_3}{\mu} \Big)$
    \vspace{1mm}    
    \State Update $J$ as: $\hspace{6pt} J = \argmin_{J} \, \Big( \frac{\alpha}{\mu} \norm[\big]{J}_{*} + \frac{1}{2} \norm[\big]{J - (D_i + \frac{T2}{\mu}}_{F}^2 \Big) $
    \vspace{1mm}
    \State Update $D_i$ as: 
   $\hspace{1pt} D_i = \Big[  2 \, \frac{\lambda}{\mu} \big( P^T X_i \> {A_i^i}^T + (\sum\limits_{j=1 \atop j \neq i}^K D_j A_i^j) {A_i^i}^T \big)  + P^T X_i \> {A_i^i}^T - E_i {A_i^i}^T $ \vspace{0.7mm}
    $\hspace{75pt} + \, J  + \frac{T_1 {A_i^i}^T - T_2}{\mu} \Big] \Big( 2(\frac{\lambda}{\mu}+1) A_i^i {A_i^i}^T + I \Big) ^{-1} $
    \vspace{1mm}
    \State Update $E_i$ as: $E_i = \argmin_{E_i} \, \Big( \frac{\beta}{\mu} \norm[\big]{E_i}_{2,1} + \frac{1}{2} \norm[\big]{E_i - (P^T X_i - D_i A_i^i + \frac{T1}{\mu})}_{F}^2 \Big) $
    \vspace{1mm}   
    \State Update $T_1,T_2,T_3$ as: 
    \NoNumber {}
    $\hspace{43pt} T_1 = T_1 + \mu (P^T X_i - D_i A_i^i - E_i)$ 
    \NoNumber {$\hspace{45pt} T_2 = T_2 + \mu (D_i - J)$} 
    \NoNumber {$\hspace{45pt} T_3 = T_3 + \mu (A_i^i - Z)$} 
    \vspace{1mm}    
    \State Update $\mu$ as: $\mu = min(\rho \mu, max_{\mu})$    
    \vspace{1mm}    
    \State Check stopping conditions as: 
    \vspace{1mm}
    \NoNumber {}
    $\hspace{10pt} \norm[\big]{D_i - J}_{\infty} < \epsilon \quad \text{and} \quad \norm[\big]{P^T X_i - D_i A_i^i - E_i}_{\infty} < \epsilon \quad \text{and} \quad \norm[\big]{A_i^i - Z}_{\infty} < \epsilon $
\EndWhile
\end{algorithmic}
\end{algorithm}
\subsection{Update of Projection Matrix $P$}
In order to solve for $P$, we keep $D$ and $A$ fixed. As a result, the objective function in Eq.\ref{eq8} is then reduced to:
\begin{gather}
J_{(P)} = \argmin_{P} \Big\{ \norm[\big]{P^T X_i - D A_i^i}_{F}^2 + \norm[\big]{P^T X_i - D_i A_i^i}_{F}^2 
+ \delta \, tr(P^T X \hat{L} X^T P) \Big\}
\nonumber \\
s.t. \quad P^T P = I
\label{eq15}
\end{gather}
First, we rewrite the objective function in a more convenient form:
\begin{gather}
J_{(P)} = \argmin_{P} \Big\{ \norm[\big]{P^T X - \hat{D} \hat{Z}}_{F}^2 + \delta \, tr(P^T X \hat{L} X^T P) \Big\} \quad
s.t. \quad P^T P = I
\label{eq16}
\end{gather}
where $ \hat{D} = \Big[ [D,D_1] , [D,D_2], \dots, [D,D_K] \Big]$ and $\hat{Z}$ is a block-diagonal matrix, whose diagonal elements are formed as $\hat{Z}_{ii} = [A_i \, ; \, A_i^i]$. Because of the orthogonal constraint $P^T P = I$, we have $\norm{P^T X - \hat{D} \hat{Z}}_{F}^2 = tr(P^T \varphi(P) P) $, where $\varphi(P) = \big( X - P \hat{D} \hat{Z} \big) \big( X - P \hat{D} \hat{Z} \big)^T $. Hence, Eq.\ref{eq16} is reformulated as:
\begin{gather}
J_{(P)} = \argmin_{P}  \> tr \Big( P^T \big( \, \varphi(P)+ \delta (X \hat{L} X^T) \big) P \Big) \quad
s.t. \quad P^T P = I
\label{eq17}
\end{gather}
To solve the above minimization, we iteratively update $P$ according to the projection matrix obtained in the previous iteration. Using singular value decomposition (SVD) technique, $[U,\Sigma,V^*] =  SVD \big( \varphi(P)+ \delta (X \hat{L} X^T) \big)$. Then, we can update $P$ as the first $l$ most important eigenvectors in $U$, \ie, $P_t = U(1:l,:)$,  where $P_t$ is the projection matrix in the $t^{th}$ iteration. To avoid big changes in $P$ and make the optimization stable, we choose to update $P$ gradually in each iteration as following:
\begin{gather}
P_t =  P_{t-1} + \gamma \Big( U(1:l,:) - P_{t-1}) \Big) 
\label{eq18}
\end{gather}
$\gamma$ is a small positive constant to control the change of $P$ in consecutive iterations.
\raggedbottom
\section{The Classification Scheme}
\label{sec:classification}
Once $D$ and $P$ are learned, they could be used to represent a query sample $x_{test}$ and find its corresponding label. The test sample is projected into the low-dimensional space and coded over $D$ by solving the following equation:
\begin{gather}
\hat{a} = \argmin_{a} \big\{ \norm[\big]{P^T x_{test} - D a}_{2}^2 + \xi \norm[\big]{a}_{1} \big\}
\label{eq19}
\end{gather}
$\xi$ is a positive scalar and the coding vector $\hat{a}$ can be written as $\hat{a} = [\hat{a}_1, \hat{a}_2, \dots \hat{a}_K]$ where $\hat{a}_i$ is the coefficient sub-vector associated with sub-dictionary $D_i$. The representation residual for the $i$th class is calculated as:
\begin{gather}
e_i = \norm[\big]{P^T x_{test} - D_i \hat{a}_i}_{2}^2 + \omega \norm[\big]{\hat{a} - m_i}_{2}^2 
\label{eq20}
\end{gather}
where $\omega$ is a preset weight. Finally, the identity of testing sample is determined by $identity(x_{test}) = \argmin_{i} \{e_i\}$.
\begin{figure}[t]
\centering
\subfloat[USPS]{\includegraphics[width=5cm,keepaspectratio]{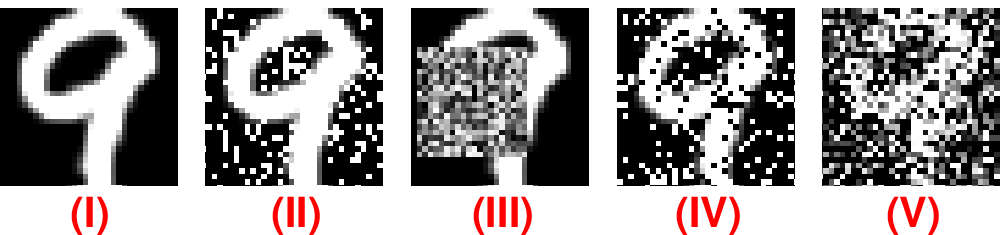}\label{fig:corruption-usps}}
\hspace{5pt} 
\subfloat[AR]{\includegraphics[width=5cm,keepaspectratio]{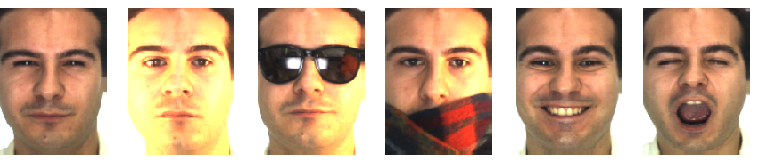}\label{fig:faces-ar}}
\vspace{-0.5cm} 
\subfloat[Extended YaleB]{\includegraphics[width=5.2cm,keepaspectratio]{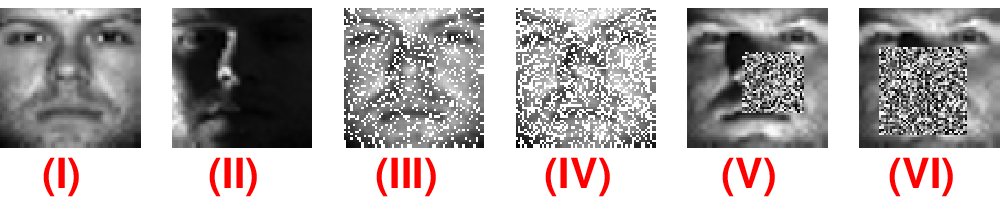}\label{fig:faces-yale}}
\hspace{4pt} 
\subfloat[COIL]{\includegraphics[width=5.2cm,keepaspectratio]{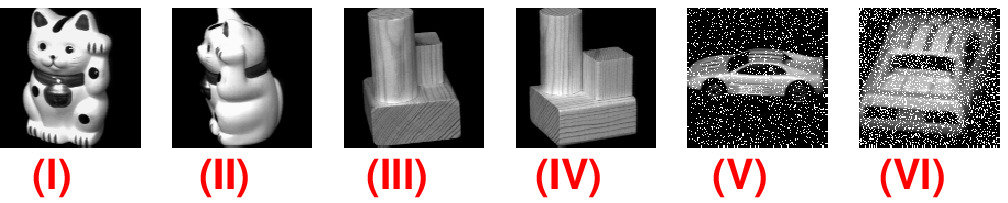}\label{fig:objects-coil}}
\vspace{-0.4em}
\caption{Sample images from \textbf{(a)}USPS dataset with (ii)pixel (iii)block (iv)salt \& pepper (v)Gaussian noise \textbf{(b)}AR dataset \textbf{(c)}Extended YaleB dataset with pixel and block corruption \textbf{(d)}COIL dataset with pixel corruption}
\label{fig:corruption-all-datasets}
\vspace{-1.5em}
\end{figure}
\raggedbottom
\section{Experimental Results}
\label{sec:experiment}
The performance of JPDL-LR method is evaluated on various image classification tasks. We compare our method with the state-of-the-art methods on the robustness to dimensionality reduction and different types of noise. Generally, in the experiments we compare our method with three types of methods: 

\noindent$-$DR methods: We compare the proposed method with several DR methods such as PCA \cite{PCA}, LDA \cite{LDA} and LPP \cite{LPP}. PCA and LDA are representative unsupervised and supervised subspace learning methods, which are optimal in the sense of reconstruction error and classification respectively. LPP considers local neighborhood structure by constructing a neighborhood graph of the training data, which makes it less sensitive to outliers \cite{DR-Survey}. These methods are followed by a multi-class linear SVM classifier.
 
\noindent$-$DL methods: We compare the results with conventional discriminative DL methods as well as the discriminative LR dictionary learning methods. FDDL \cite{FDDL} introduces Fisher criterion on the coding vectors to enhance class discrimination; whereas, D\textsuperscript{2}L\textsuperscript{2}R\textsuperscript{2} \cite{D2L2R2} adopts Fisher discrimination and meantime imposes a LR constraint on sub-dictionaries to make them robust to noise and achieves impressive results especially when corruption existed. PCA (\ie \, Eigenface) or random projection \cite {RP} (\ie \, random face) is used for dimensionality reduction of features before the dictionary learning process. 

\noindent$-$Joint DR and DL methods: These methods generally share the same idea of formulating the projection and dictionary learning into a unified optimization framework. We compare our method with DR-SRC \cite{DR-SRC}, SE \cite{SE}, LGE-KSVD \cite{LGE-KSVD}, JDDRDL \cite{JDDRDL} and JNPDL \cite{JNPDL} which already introduced. Note that since SE can obtain at most $K$ (\ie \, number of classes) features in the reduced space, it would be excluded from the experiment which is not applicable.
\begin{figure}[t]
\centering
\subfloat[]{\includegraphics[width=4cm,keepaspectratio]{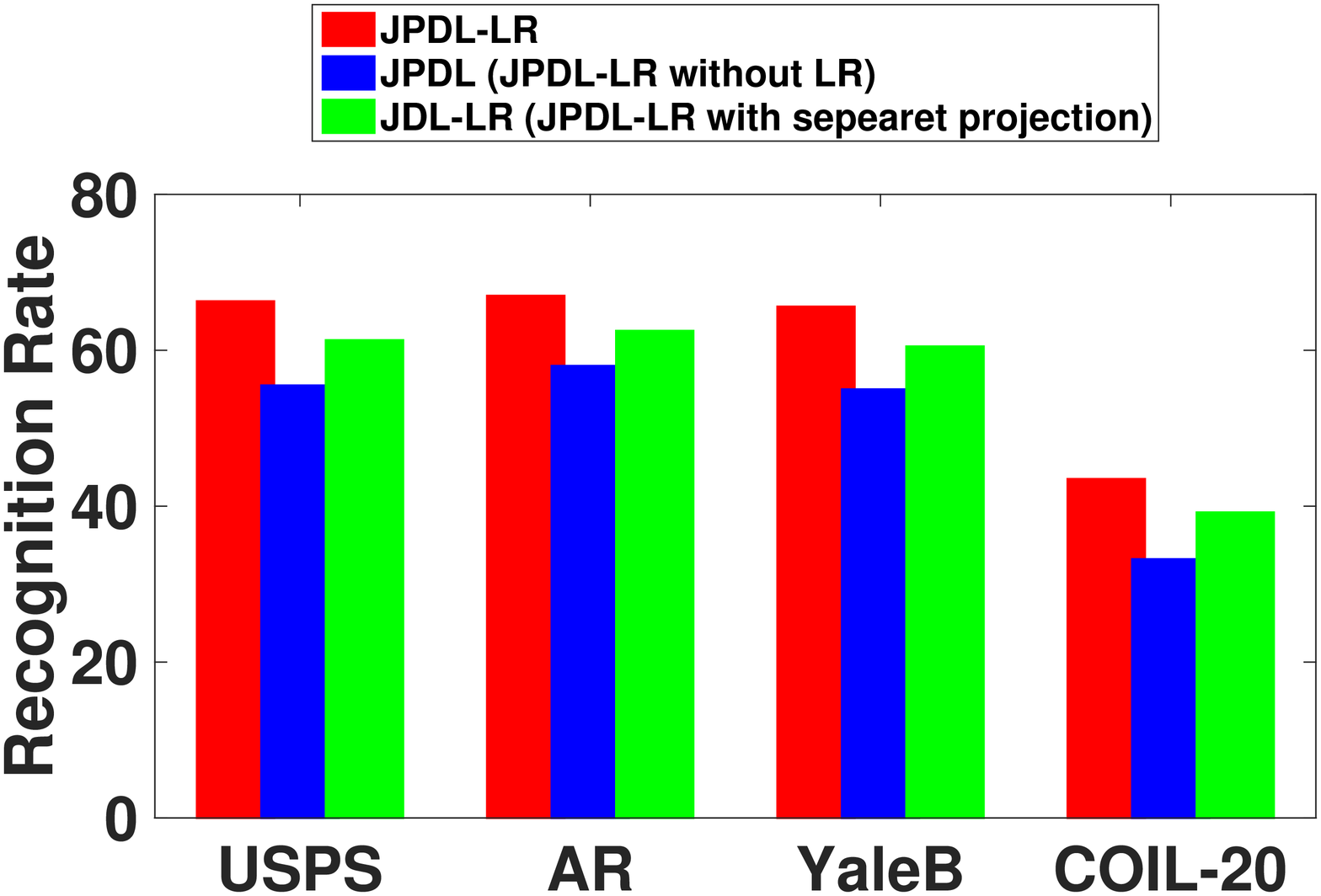}\label{fig:components}}
\hspace{3pt} 
\subfloat[]{\includegraphics[width=3.1cm,keepaspectratio]{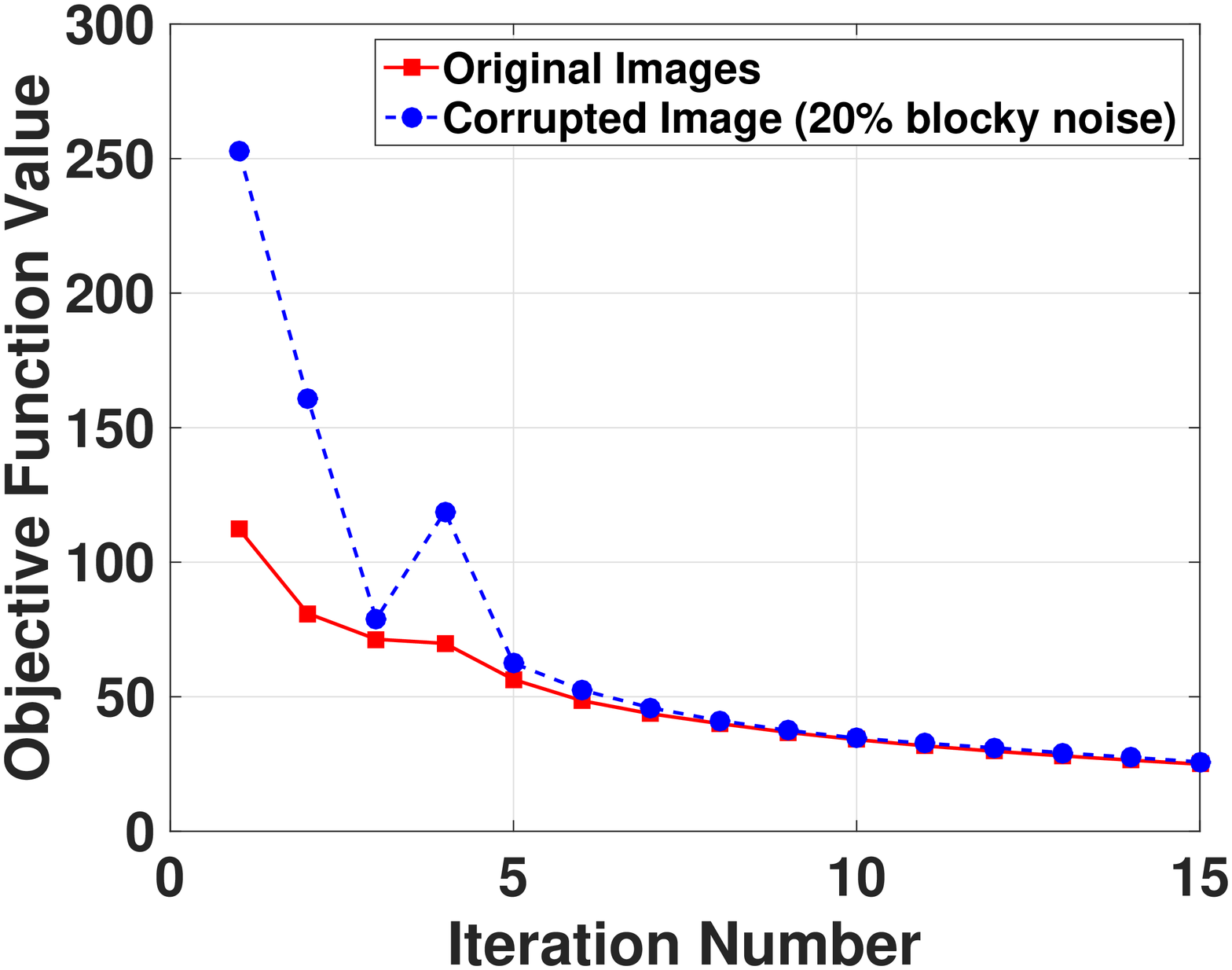}\label{fig:convergence}}
\hspace{3pt} 
\subfloat[]{\includegraphics[width=3.9cm,keepaspectratio]{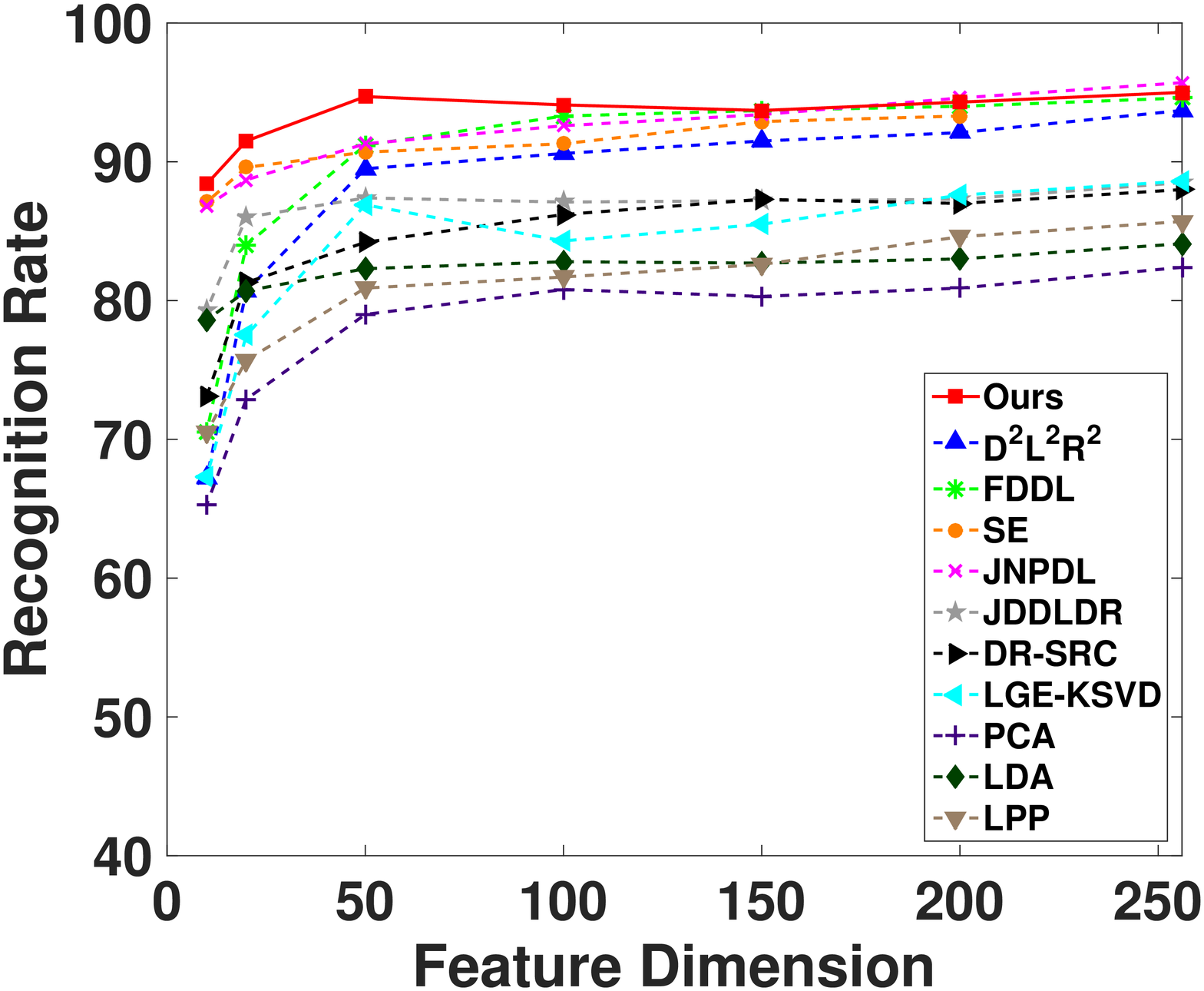}\label{fig:usps-noiseless}}
\vspace{-1em}
\caption{\textbf{(a)}Efficacy of different components of JPDL-LR under $d=0.1m$ and 40\% pixel corruption\textbf{(b)}Convergence curve of JPDL-LR method on USPS dataset \textbf{(c)}Accuracy versus feature dimension on noiseless USPS dataset}
\vspace{-1.5em}
\end{figure}
\subsection{Parameter Selection}
There are nine parameters, which need to be tuned in our method: $\lambda_1$, $\lambda_2$,$\alpha$, $\eta$, $\delta$ in Eq.\ref{eq8}; $\beta$, $\lambda$ in Eq.\ref{eq13}; $\xi$ in Eq.\ref{eq19} and $\omega$ in Eq.\ref{eq20}. We found out that changing $\eta$, $\alpha$, $\lambda$ and $\omega$ would not affect the results that much, and we set them as $1,1,1,0.001$ respectively. In all experiments, the other tuning parameters of JPDL-LR and all the competing methods are chosen by 5-fold cross validation. Generally, we select images randomly for constructing training set and the random selection process is repeated $10$ times and we report the average recognition rates for all methods. Also, we set the maximum iteration of iterative methods as $10$. 
\subsection{Digit Recognition}
We evaluate the performance of our method on the USPS \cite{USPS} handwritten digit dataset, which has 7,291 training and 2,007 test images, each of size $16\times 16$. To test the robustness to noise, we simulate different types of noise in this experiment, including Gaussian noise, salt \& pepper noise, pixel corruption and block corruption. For pixel corruption, we replace a certain percentage (from 10\% to 50\%) of randomly selected pixels of each image with pixel value 255. For block corruption, the images are manually corrupted by an unrelated block image at a random location and the percentage of corrupted area is increased from 10\% to 50\%. 
\begin{figure}[t]
\centering
\subfloat{\includegraphics[width=3cm,keepaspectratio]{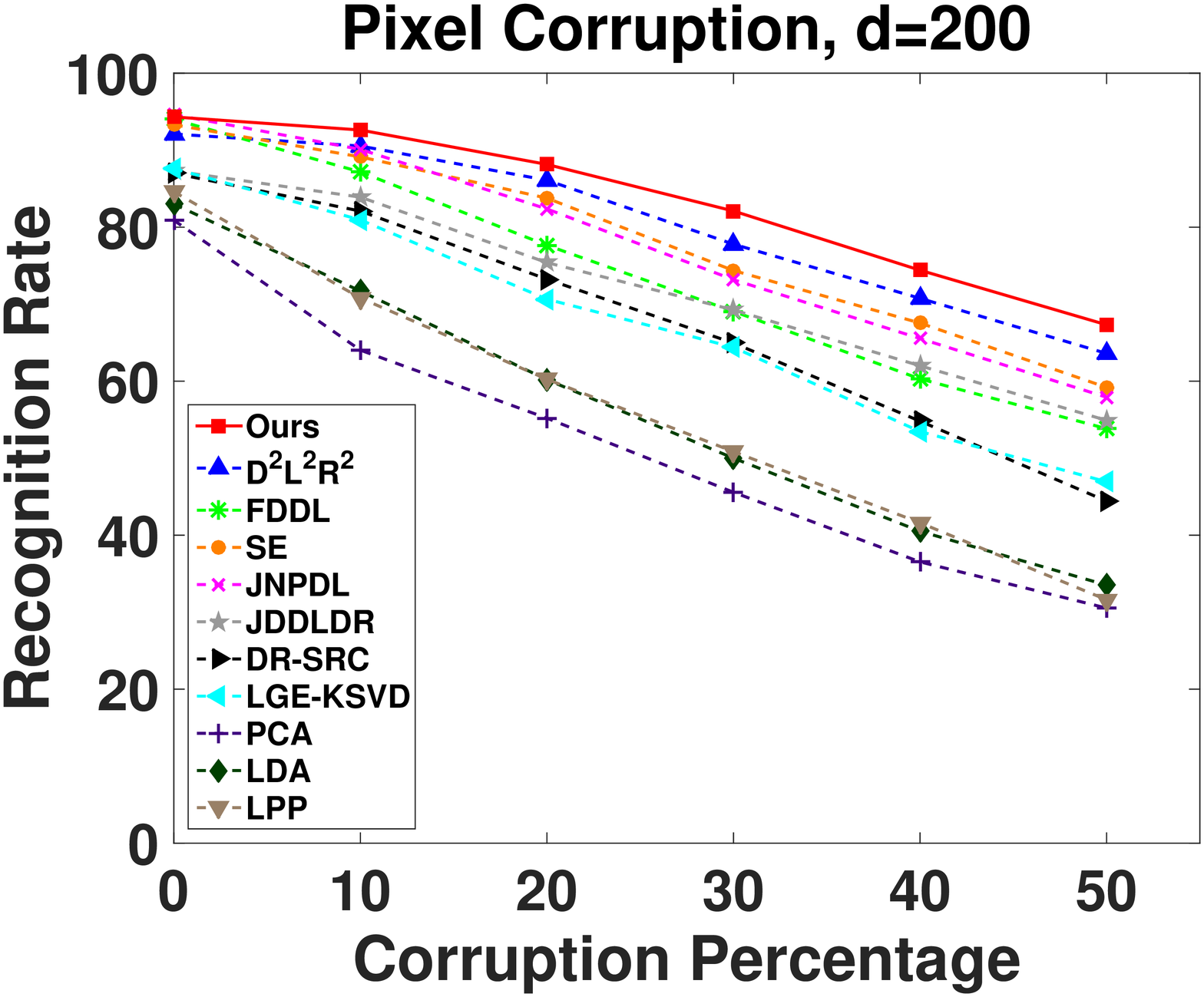}}   
\subfloat{\includegraphics[width=3cm,keepaspectratio]{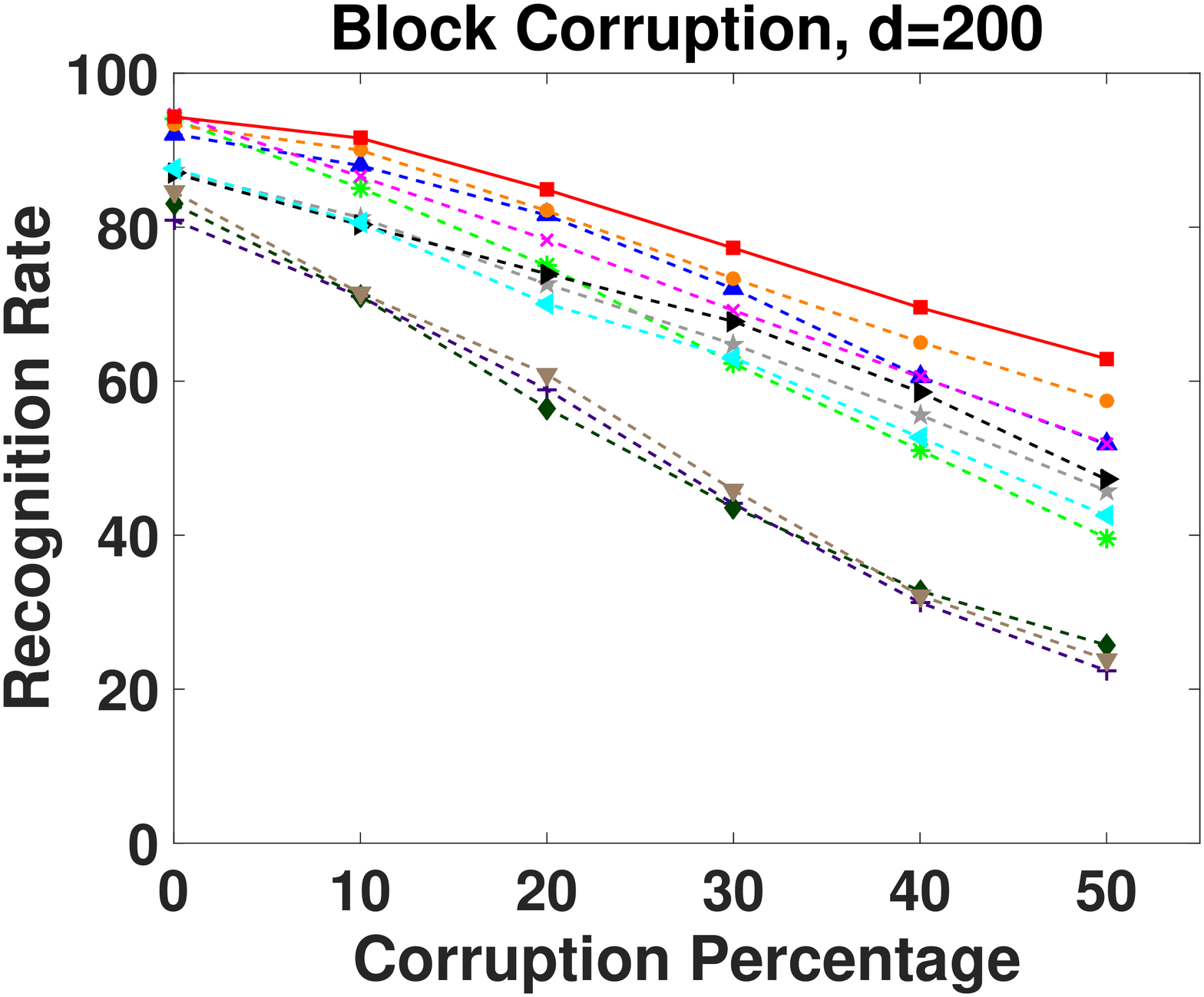}}   
\subfloat{\includegraphics[width=3cm,keepaspectratio]{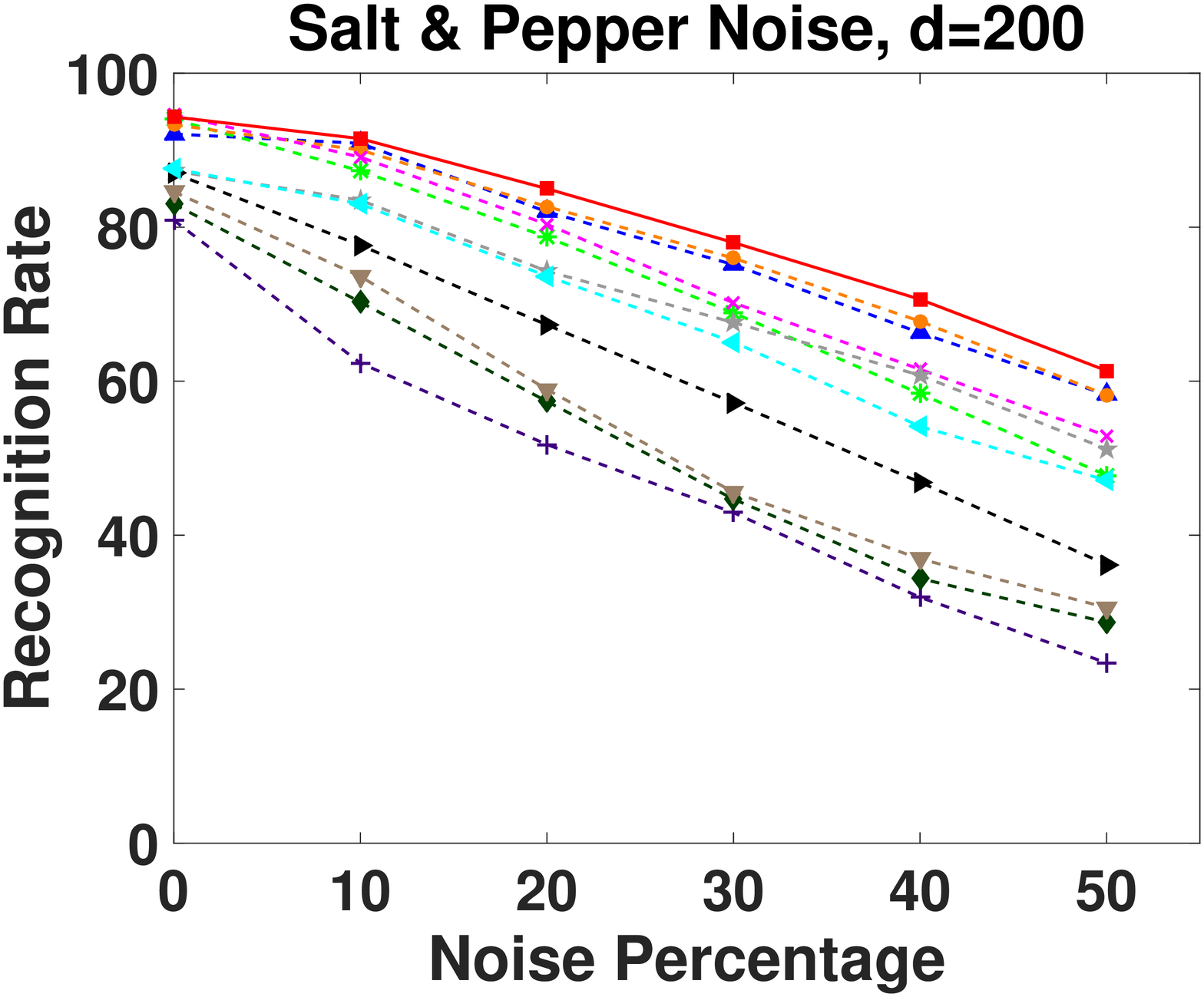}}      
\subfloat{\includegraphics[width=3cm,keepaspectratio]{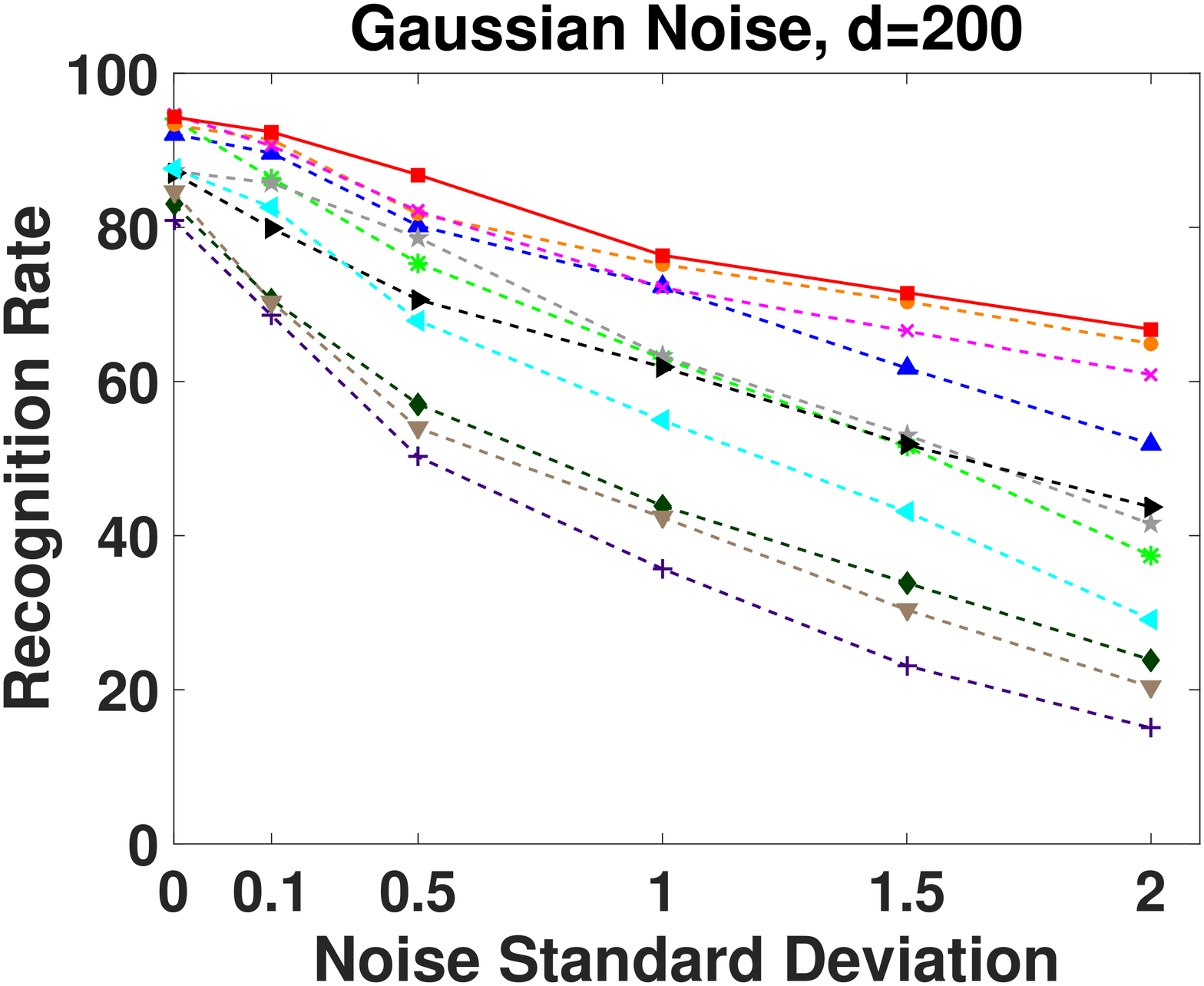}}
\\
\subfloat{\includegraphics[width=3cm,keepaspectratio]{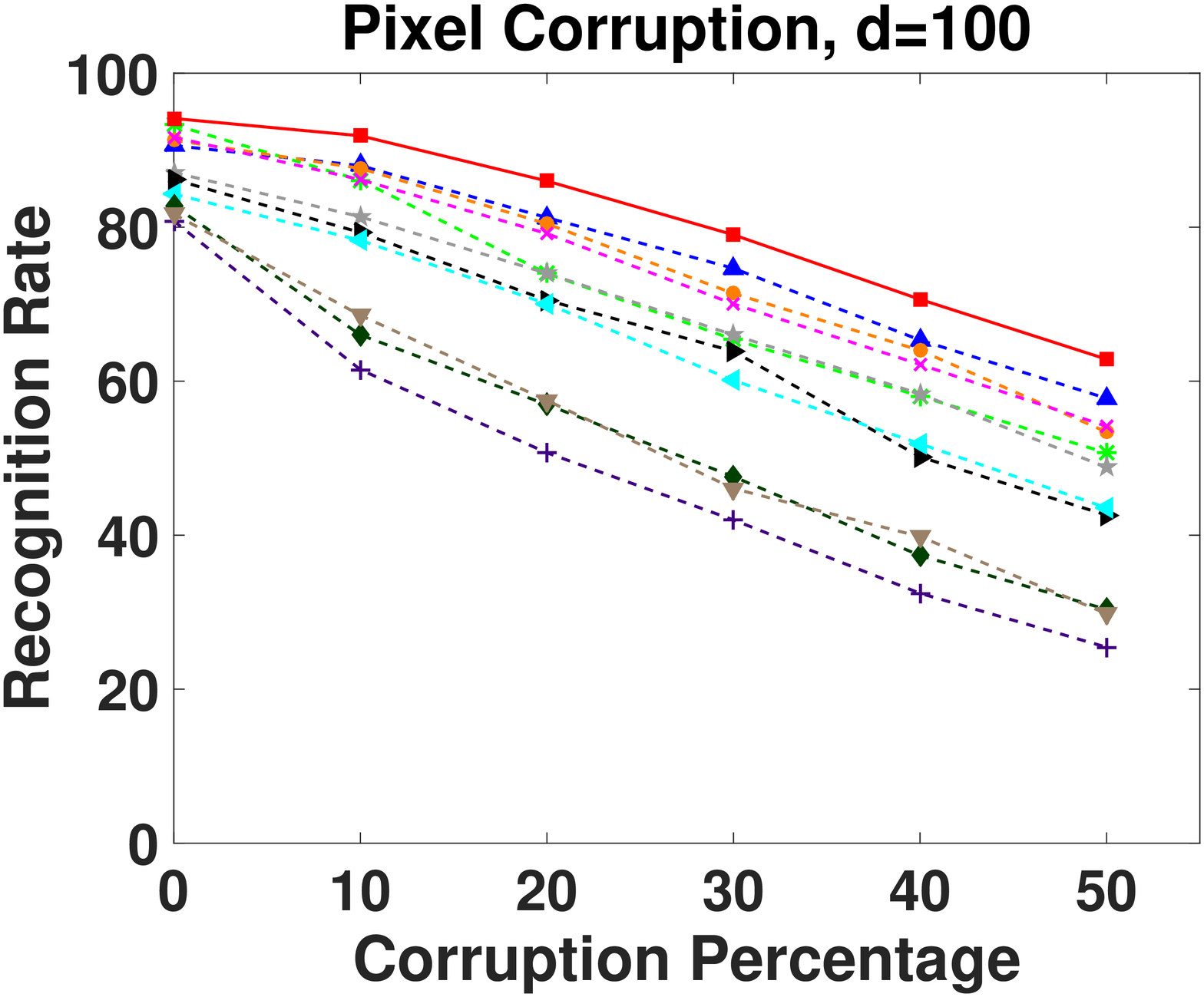}}   
\subfloat{\includegraphics[width=3cm,keepaspectratio]{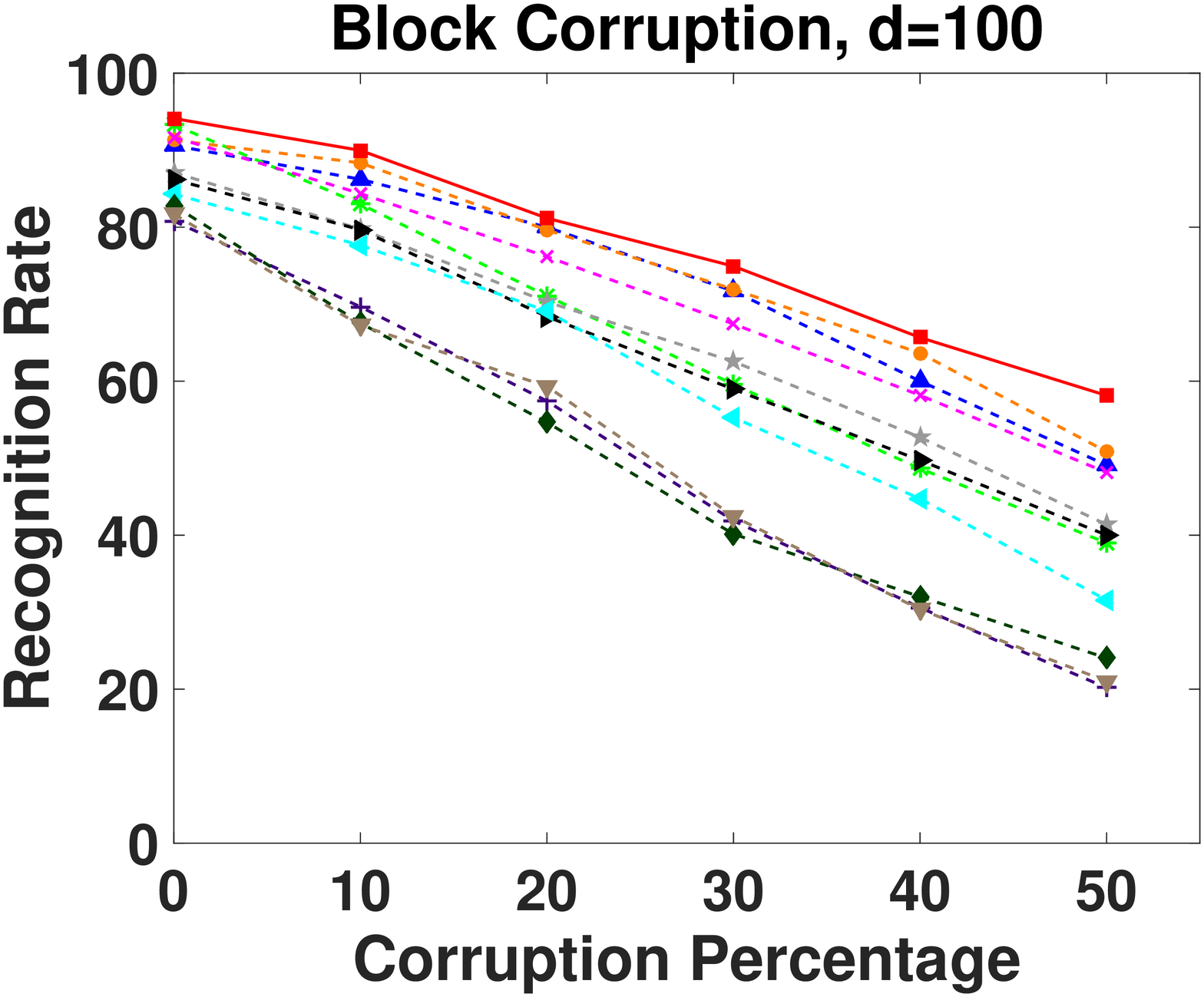}}   
\subfloat{\includegraphics[width=3cm,keepaspectratio]{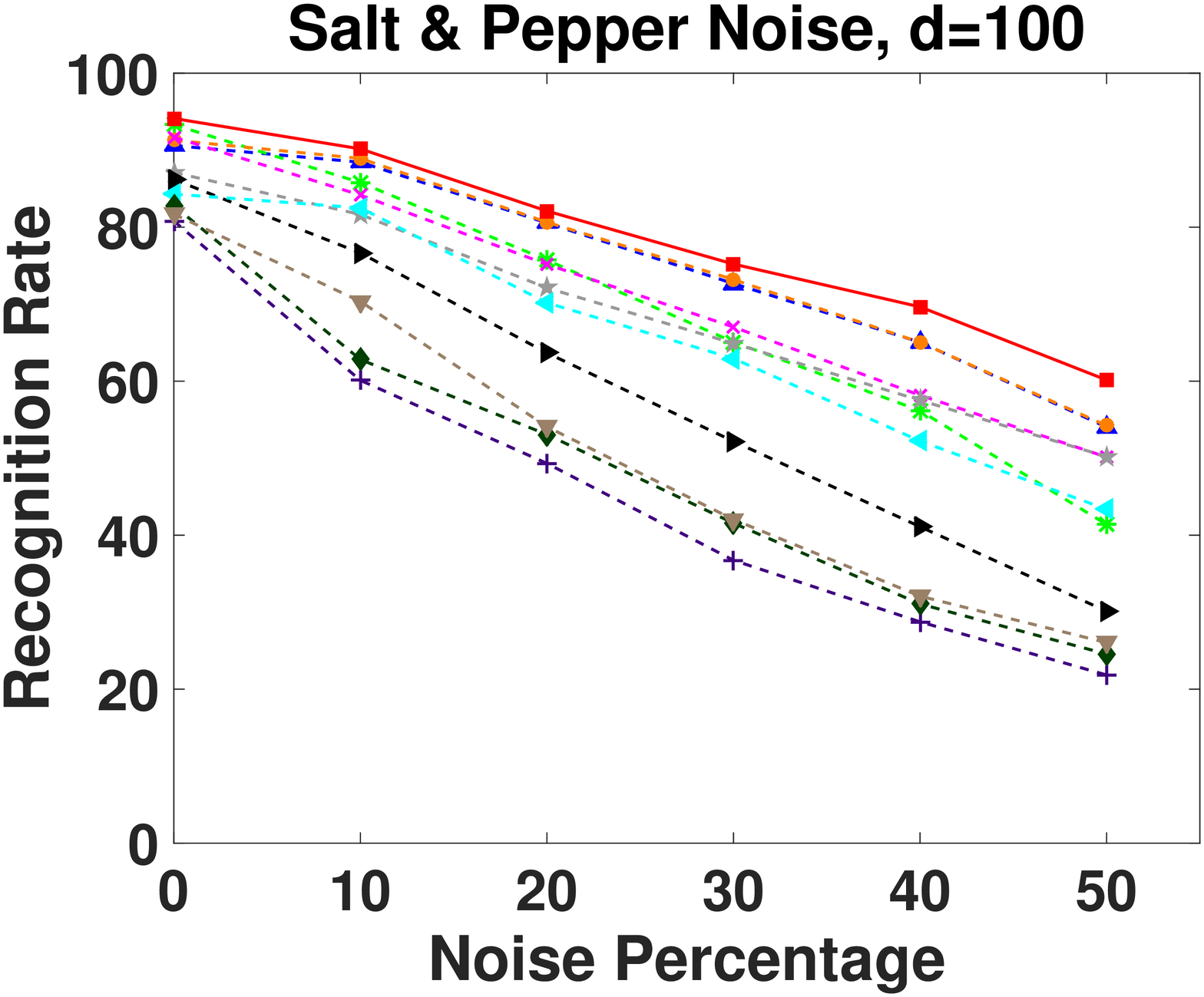}}      
\subfloat{\includegraphics[width=3cm,keepaspectratio]{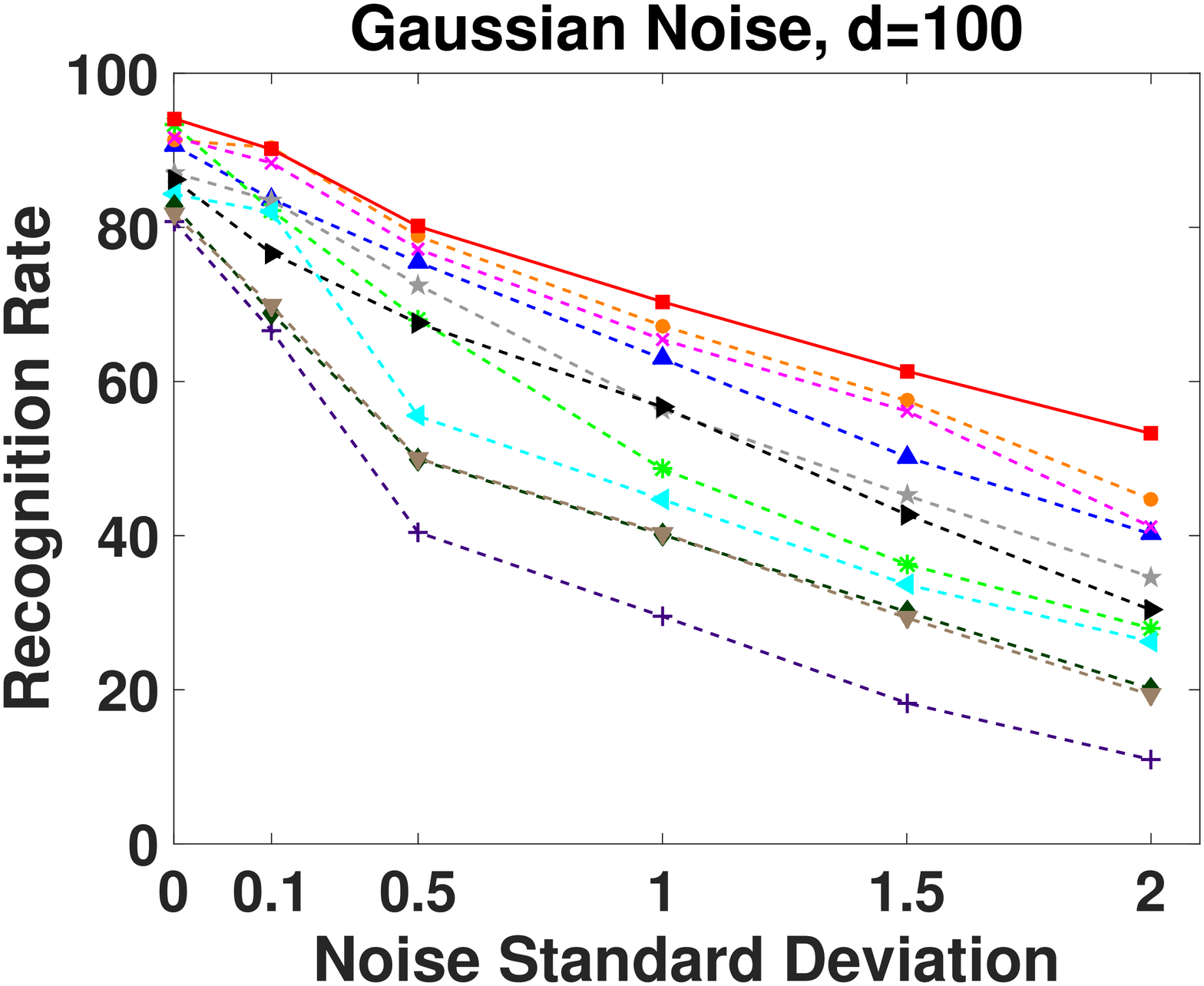}}
\\ 
\subfloat{\includegraphics[width=3cm,keepaspectratio]{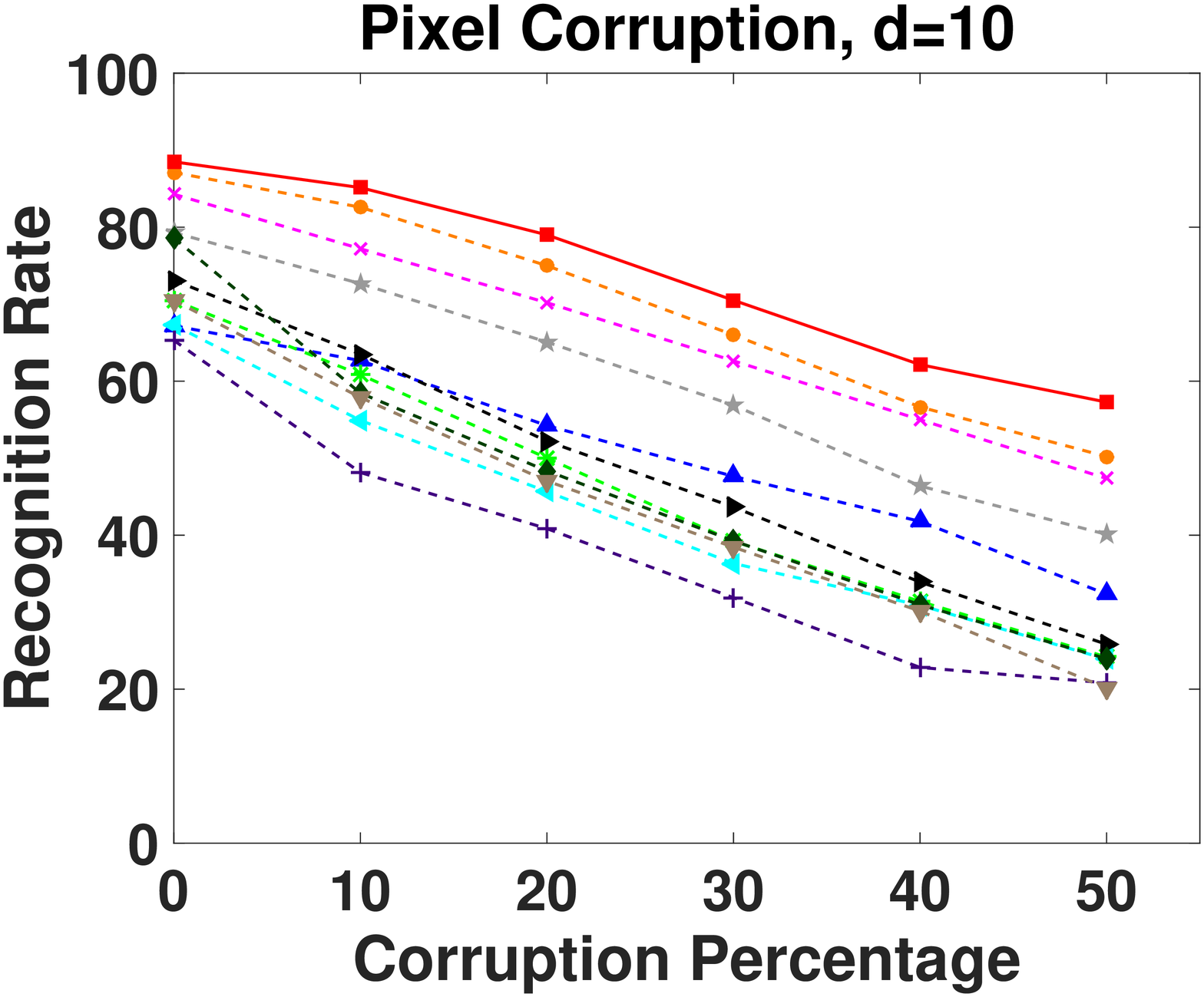}}    
\subfloat{\includegraphics[width=3cm,keepaspectratio]{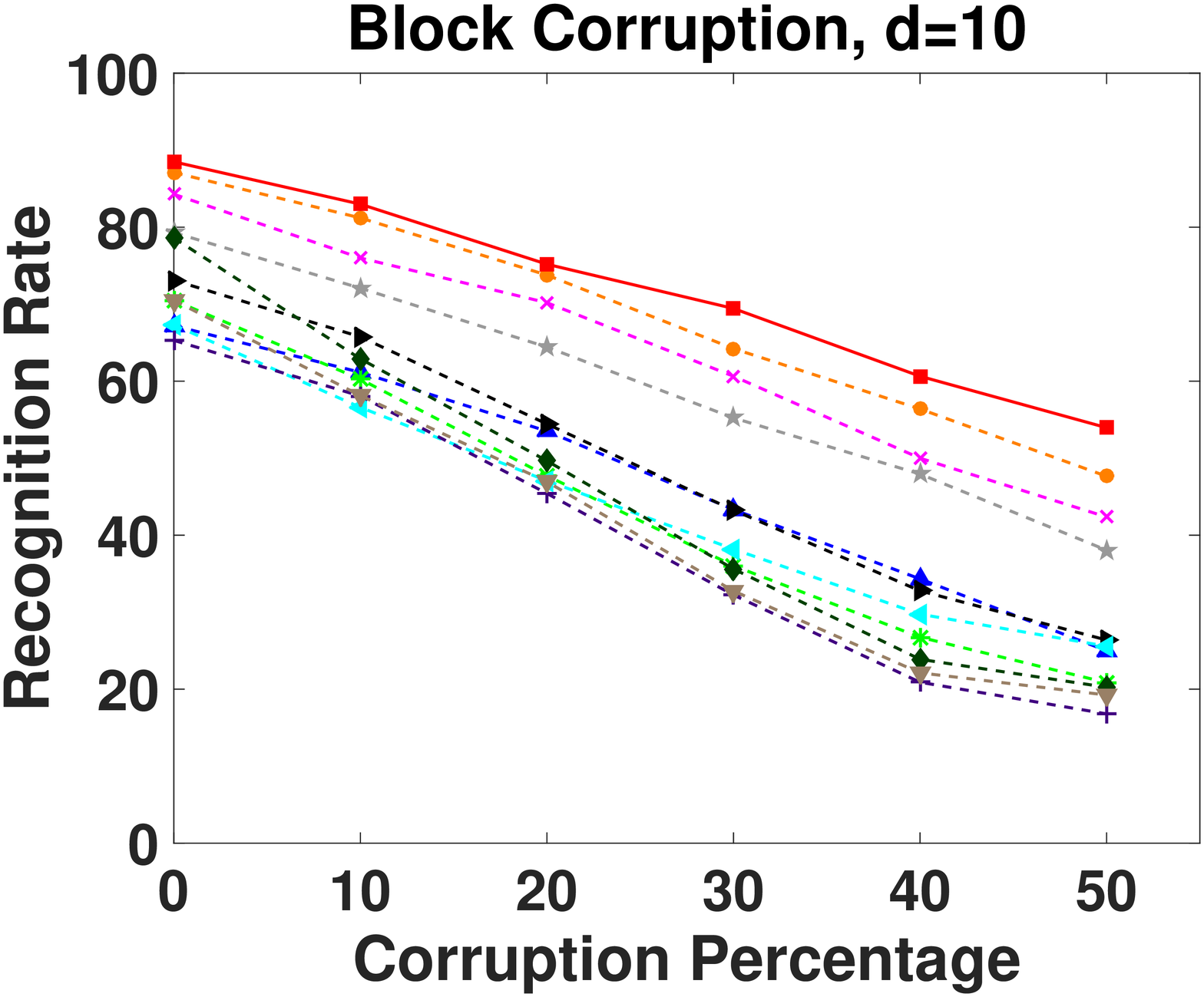}}    
\subfloat{\includegraphics[width=3cm,keepaspectratio]{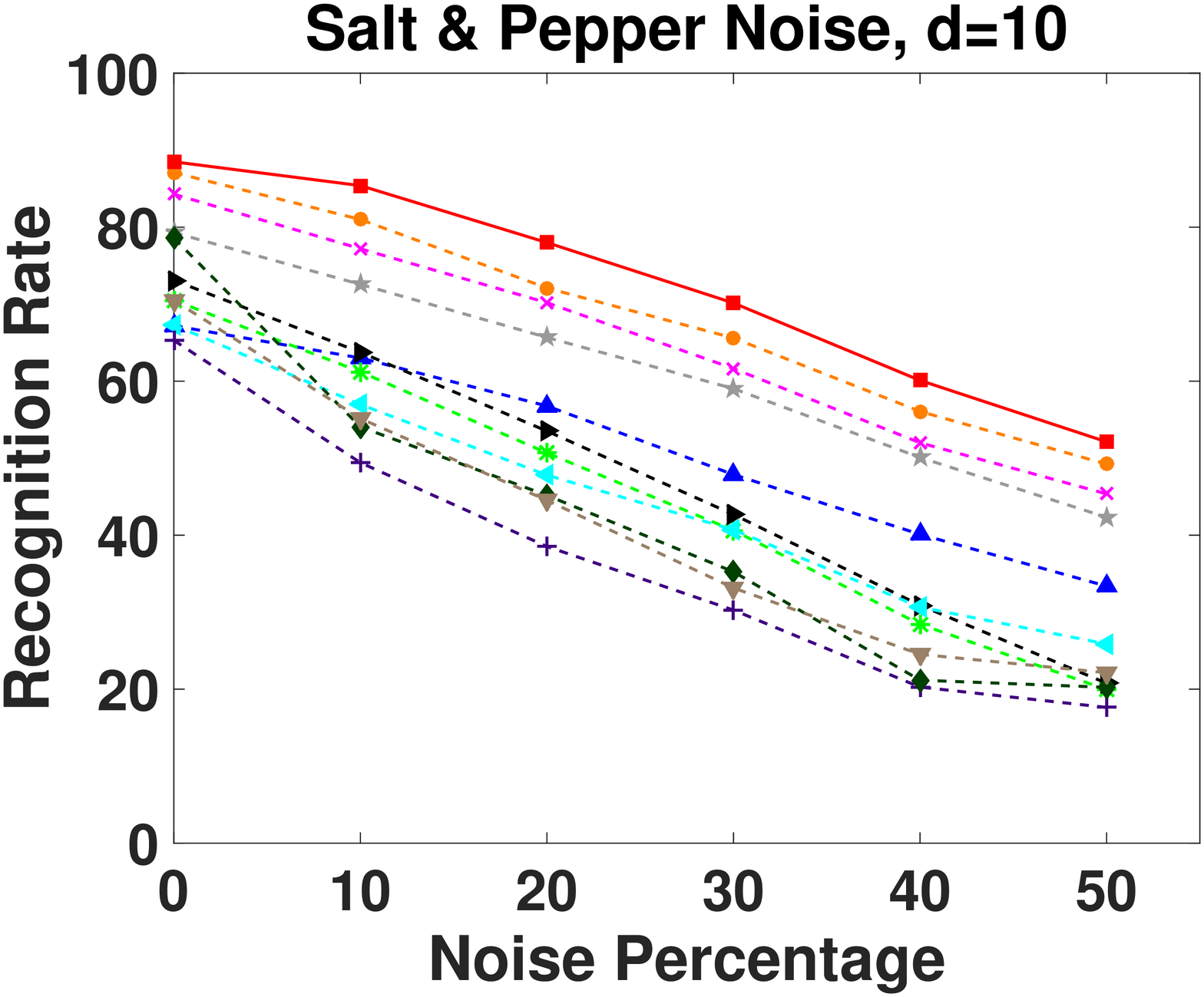}}        
\subfloat{\includegraphics[width=3cm,keepaspectratio]{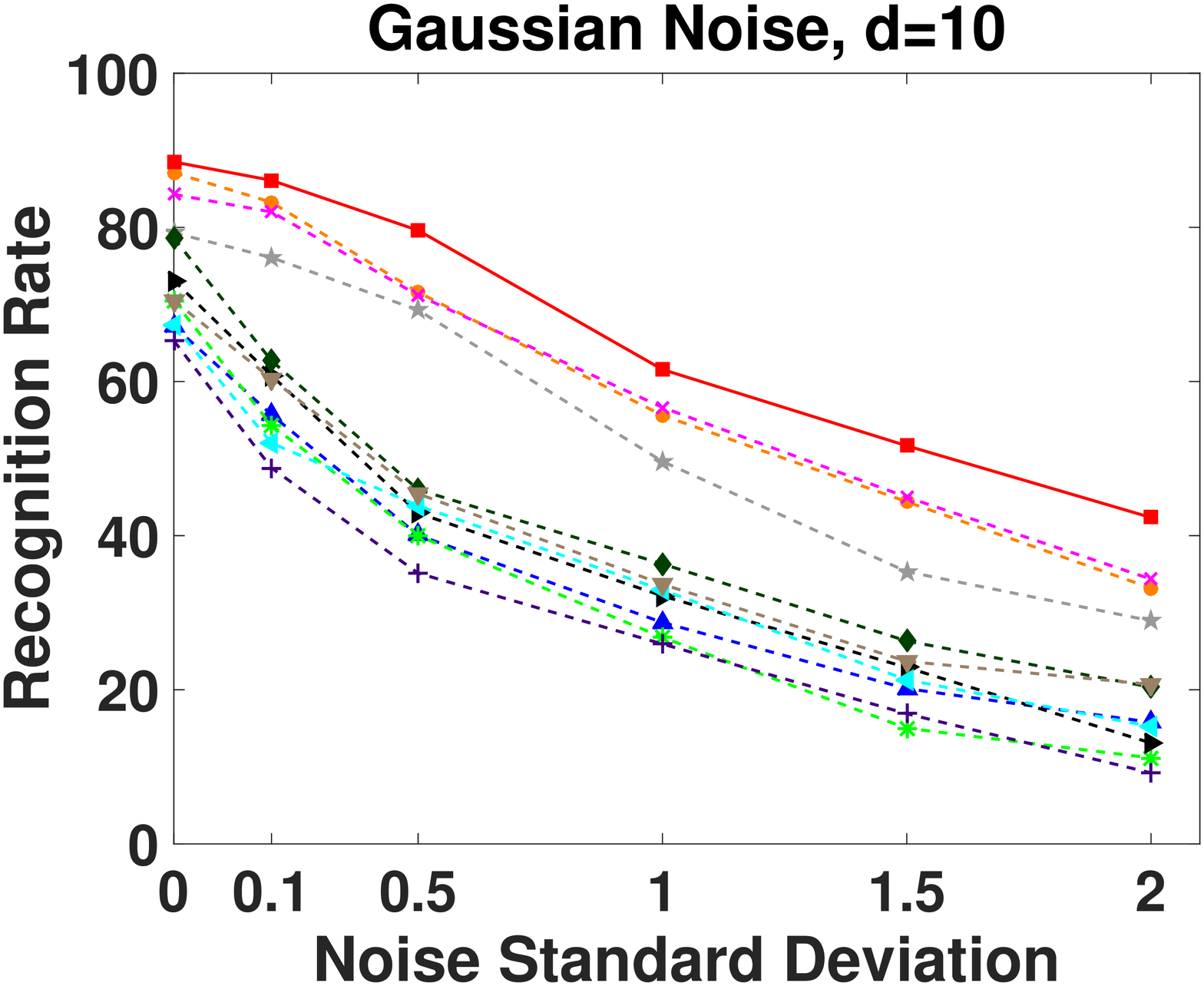}}   
\caption{Accuracy against various types of noise using different feature dimensions}
\label{fig:usps-noisy}
\vspace{-1.5em}
\end{figure}
In addition, samples are corrupted by Gaussian noise with zero mean and different standard deviations. Fig.~\ref{fig:corruption-usps} shows several examples of images with different types of noise on USPS dataset. 
In this experiment, the raw images are directly used as the features and the number of atoms in each sub-dictionary is set to 200. Fig.~\ref{fig:convergence} illustrates the convergence curves of JPDL-LR method on the original and corrupted images of USPS dataset. It can be observed that the objective function value on corrupted images (20\% blocky noise) is larger than that of original ones; however, the function converges very well after some iterations in both cases. 

Fig.~\ref{fig:usps-noiseless} shows the recognition rates of JPDL-LR and compared methods versus feature dimension on USPS dataset without any noise. We observe our JPDL-LR method is superior or competitive to other methods across different dimensions and maintains a relatively stable performance in lower dimensions. As the dimensionality decreases, the performance of FDDL along with D\textsuperscript{2}L\textsuperscript{2}R\textsuperscript{2} drops rapidly. However, joint DR-DL methods better preserve the discriminative information in relatively low dimensions compared to pre-learned ones such as PCA or random projection. When the images contain noise, the recognition rates of DR methods are severely degraded. Then, we investigate the robustness of our method versus different types of noise and evaluate the recognition rate under different dimensions when training and test samples are contaminated. We conduct experiments with three projected dimensions equal to $10$, $100$ and $200$ under different levels of four types of noise. Fig.~\ref{fig:usps-noisy} demonstrates the recognition rates of all compared methods versus varying feature dimensions on corrupted USPS dataset. In this figure, each row consists of four aforementioned types of noise for a specific dimension. 

We observe that our JPDL-LR method consistently outperforms the compared methods under different levels of corruption of various types of noise across all dimensions. We also note that, D\textsuperscript{2}L\textsuperscript{2}R\textsuperscript{2} can obtain good performance in larger dimensions ($d=200$) of heavily corrupted images; however, random projection, which is used here for DR of DL methods, fails to preserve the discriminative information in low dimensions ($d=10$). FDDL suffers from both weakness of random projection and non-robustness toward noise. Although joint DR-DL methods and in particular JNPDL perform well in noiseless observations, they easily fail to handle the large noise. Equally important JNPDL, JDDRDL and SE generally can preserve discriminative information even in relatively low feature dimensions. When images are highly contaminated and dimension is very low, the performance difference between JPDL-LR and other methods is significant. 
\subsection{Face Recognition}
\begin{figure}[t]
\centering
\subfloat[]{\includegraphics[width=3.7cm,keepaspectratio]{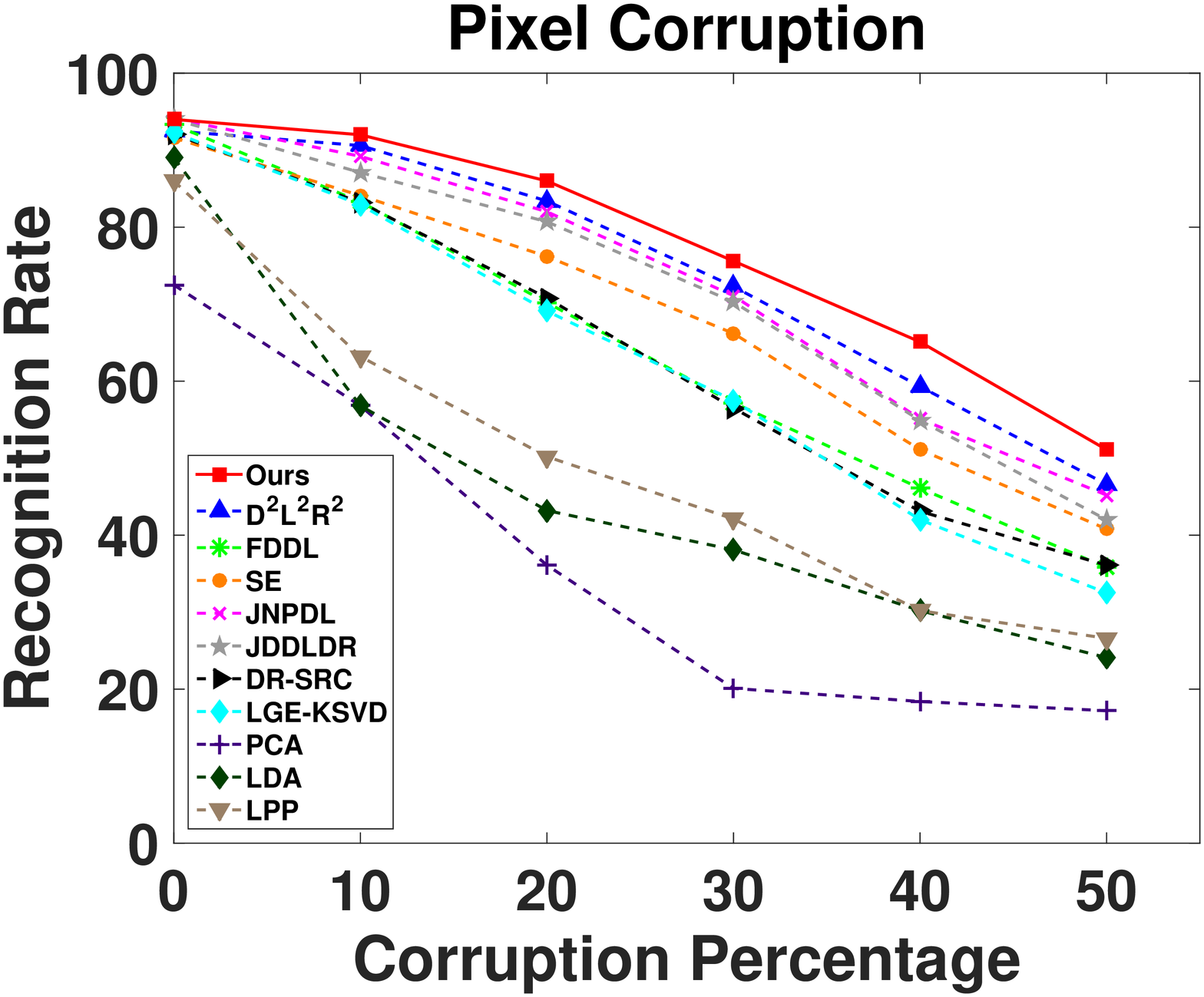} \label{fig:Yale-pixel}}  
\subfloat[]{\includegraphics[width=3.7cm,keepaspectratio]{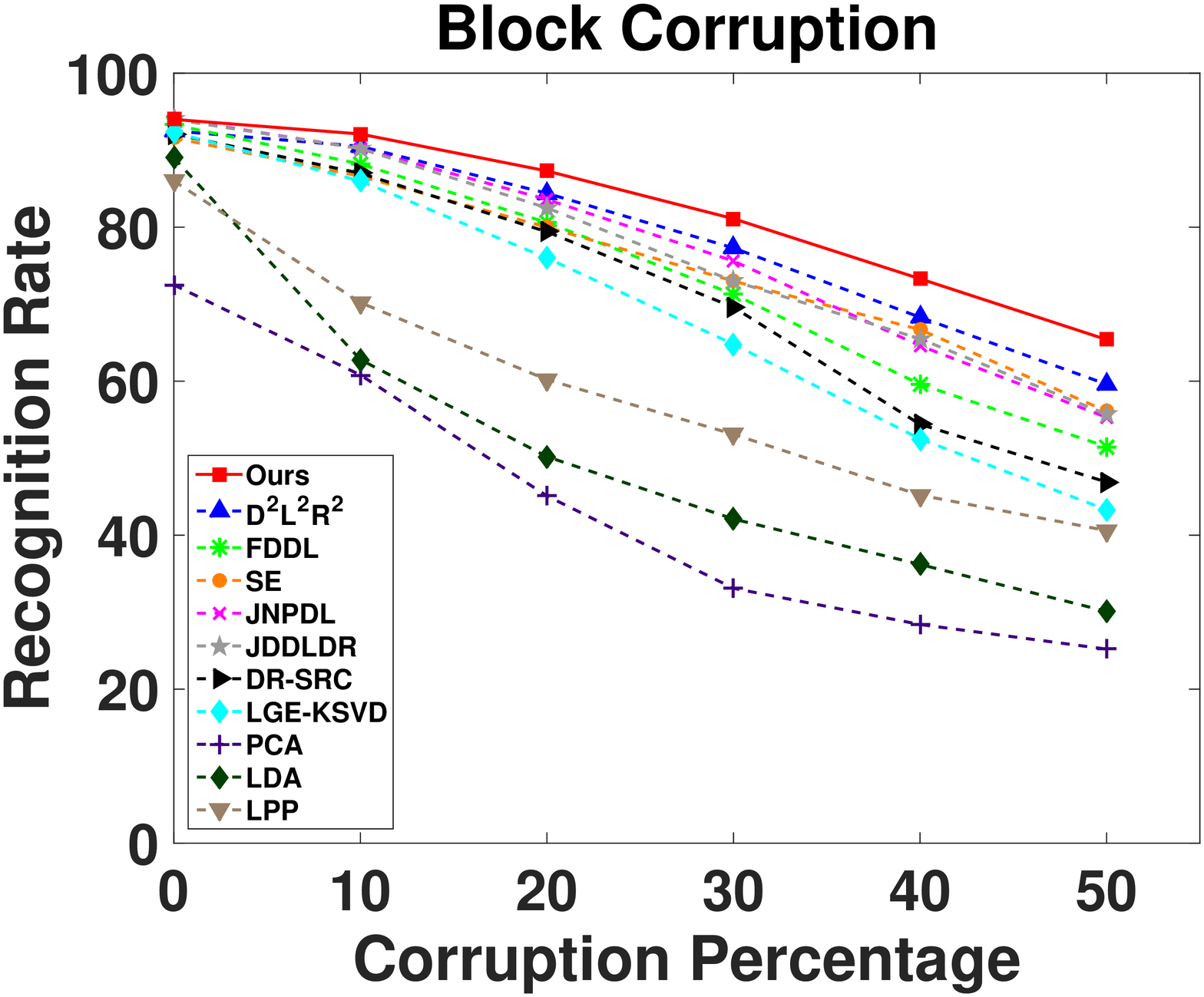} \label{fig:Yale-block}}  
\subfloat[]{\includegraphics[width=4.5cm,keepaspectratio]{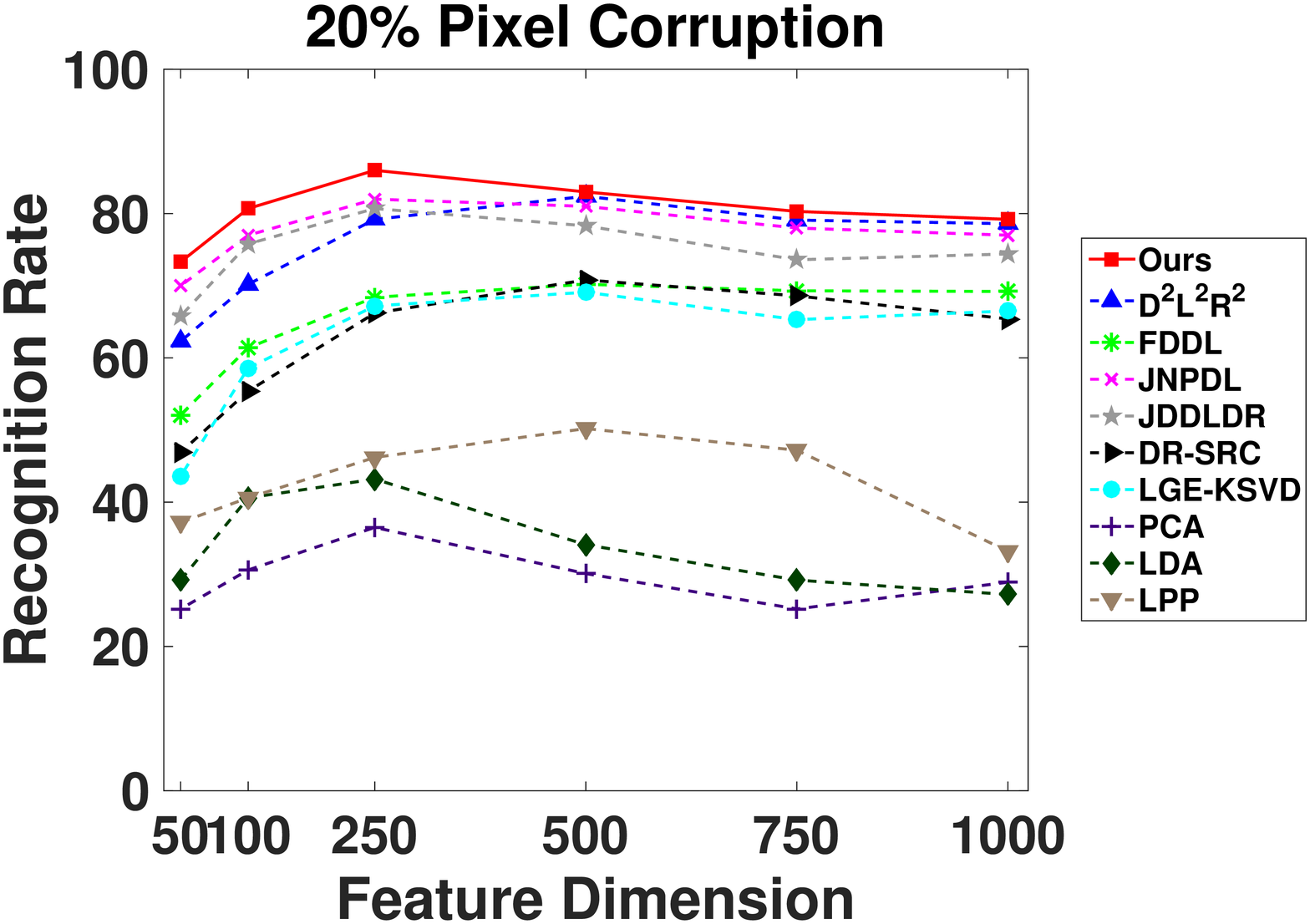}  \label{fig:Yale-dim-noise}}  
\vspace{-1em}
\caption{Recognition rates of all compared methods on YaleB dataset \textbf{(a)}under pixel corruption \textbf{(b)}under block corruption \textbf{(c)}versus feature dimension}
\vspace{-1.5em}
\end{figure}

\begin{enumerate} [label=(\Alph*)] 
 \item \textbf{AR Dataset:} The AR face dataset \cite{AR} includes over $4,000$ frontal face images from $126$ individuals. We select a subset of $2,600$ images from $50$ male and $50$ female subjects in the experiments. In each session, each person has 13 images, of which 3 are obscured by scarves, 3 by sunglasses and the remaining ones are of different facial expressions or illumination variations which we refer to as unobscured images. Fig.~\ref{fig:faces-ar} shows several samples of AR dataset. Each face image is resized to $27 \times 20$ and following the protocol in \cite{Structured-LR-DL}, experiments are conducted under three different scenarios:

 $-$\textbf{Sunglasses:} We select 7 unobscured images and 1 image with sunglasses from the first session as training samples for each person. The rest of unobscured images from the second session and the rest of images with sunglasses are used for testing. Sunglasses occlude about 20\% of the face image.

 $-$\textbf{Scarf:} We choose 8 training images (7 unobscured and 1 with scarf) from the first session for training, and 12 test images including 7 unobscured images from the second session and the remaining 5 images with scarf from two sessions for testing. The scarf covers around 40\% of face image.

 $-$\textbf{Sunglasses+Scarf:} We consider the case in which both training and test images are occluded by sunglasses and scarf. We select 7 unobscured, plus 2 occluded images (1 with sunglasses, 1 by scarf) from the first session for training and the remaining 17 images in two sessions for testing per class.
\begin{figure}[t]
\centering
\subfloat[]{\includegraphics[width=3.7cm,keepaspectratio]{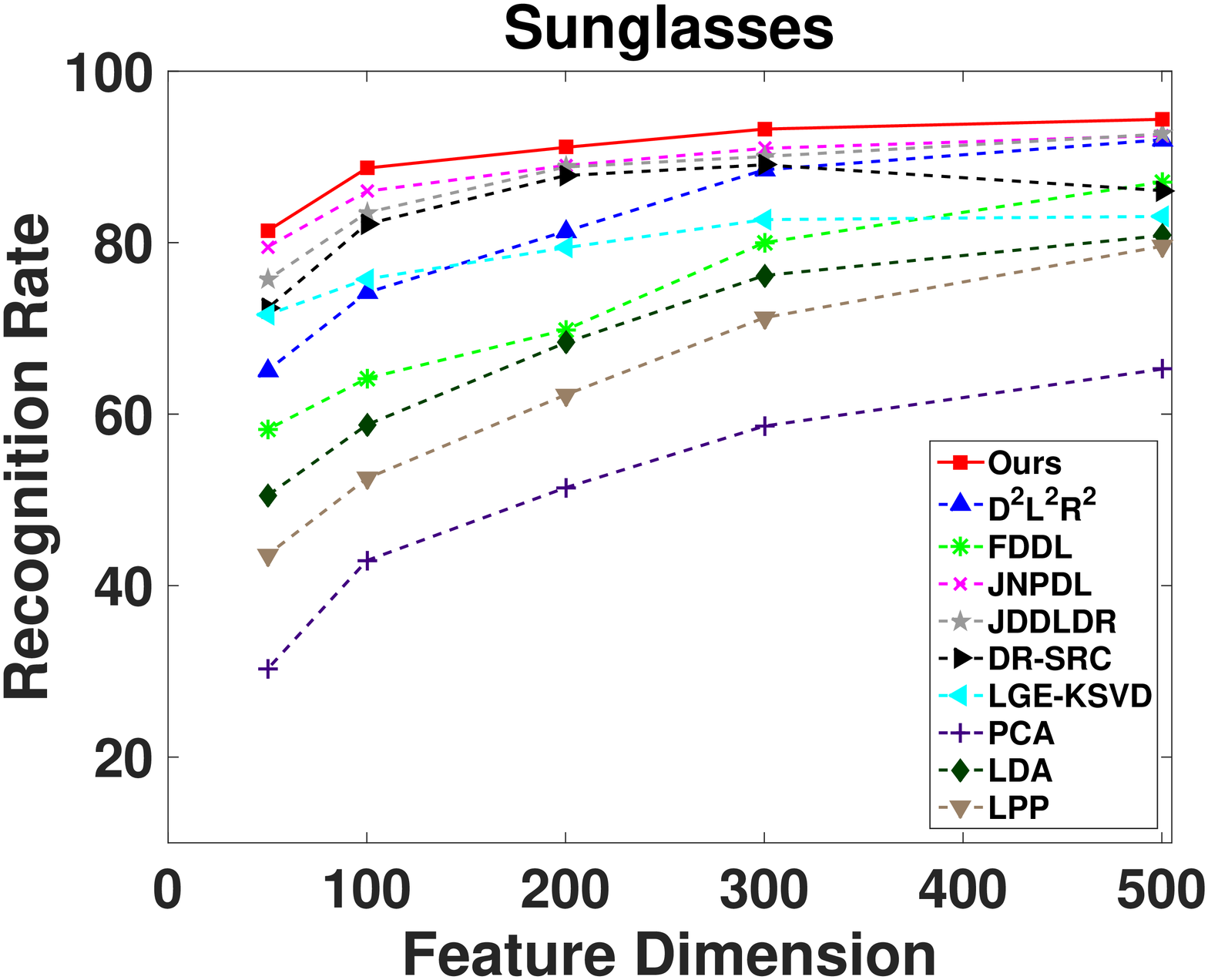} \label{fig:AR-sunglasses}}  
\subfloat[]{\includegraphics[width=3.7cm,keepaspectratio]{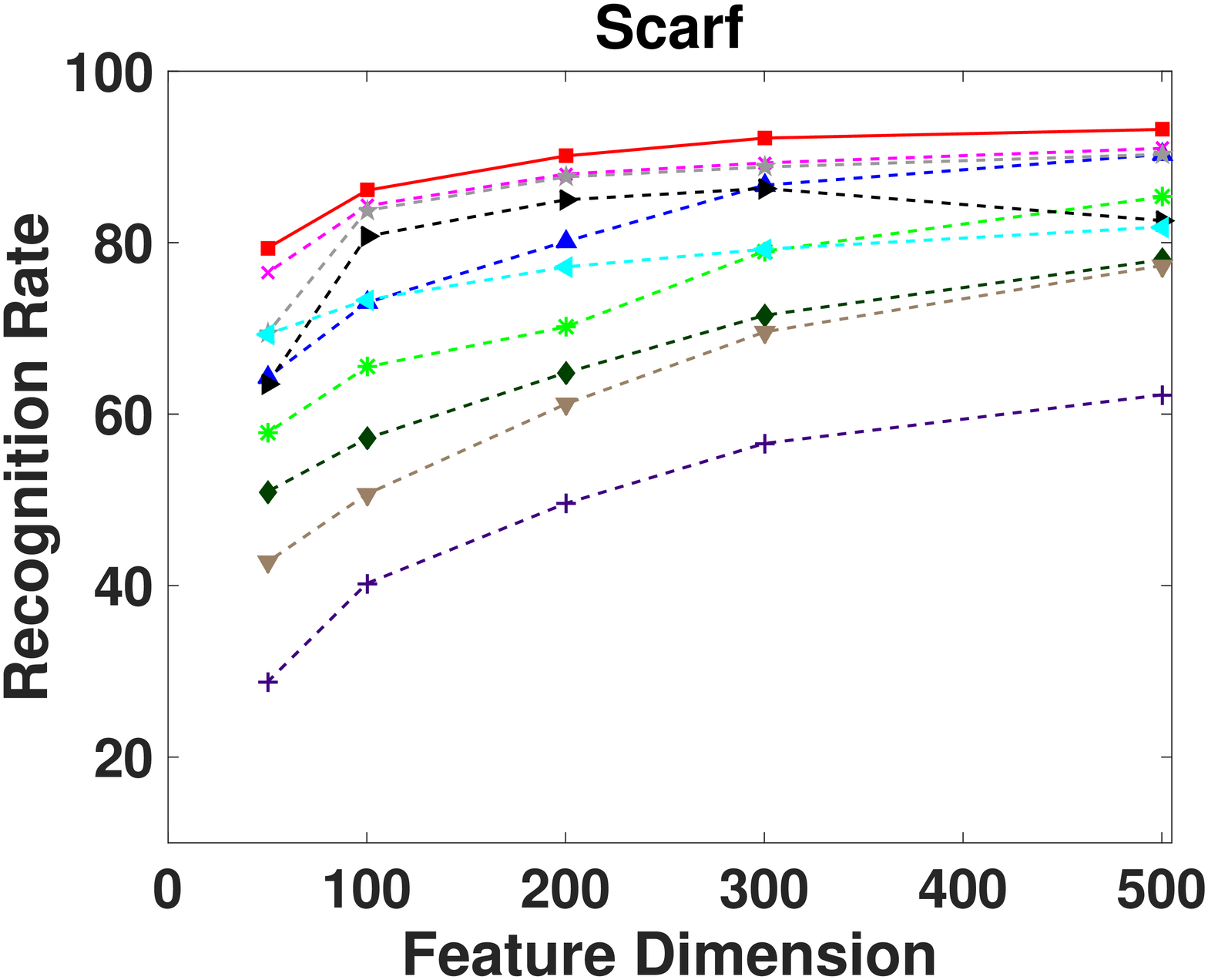} \label{fig:AR-scarf}}  
\subfloat[]{\includegraphics[width=3.7cm,keepaspectratio]{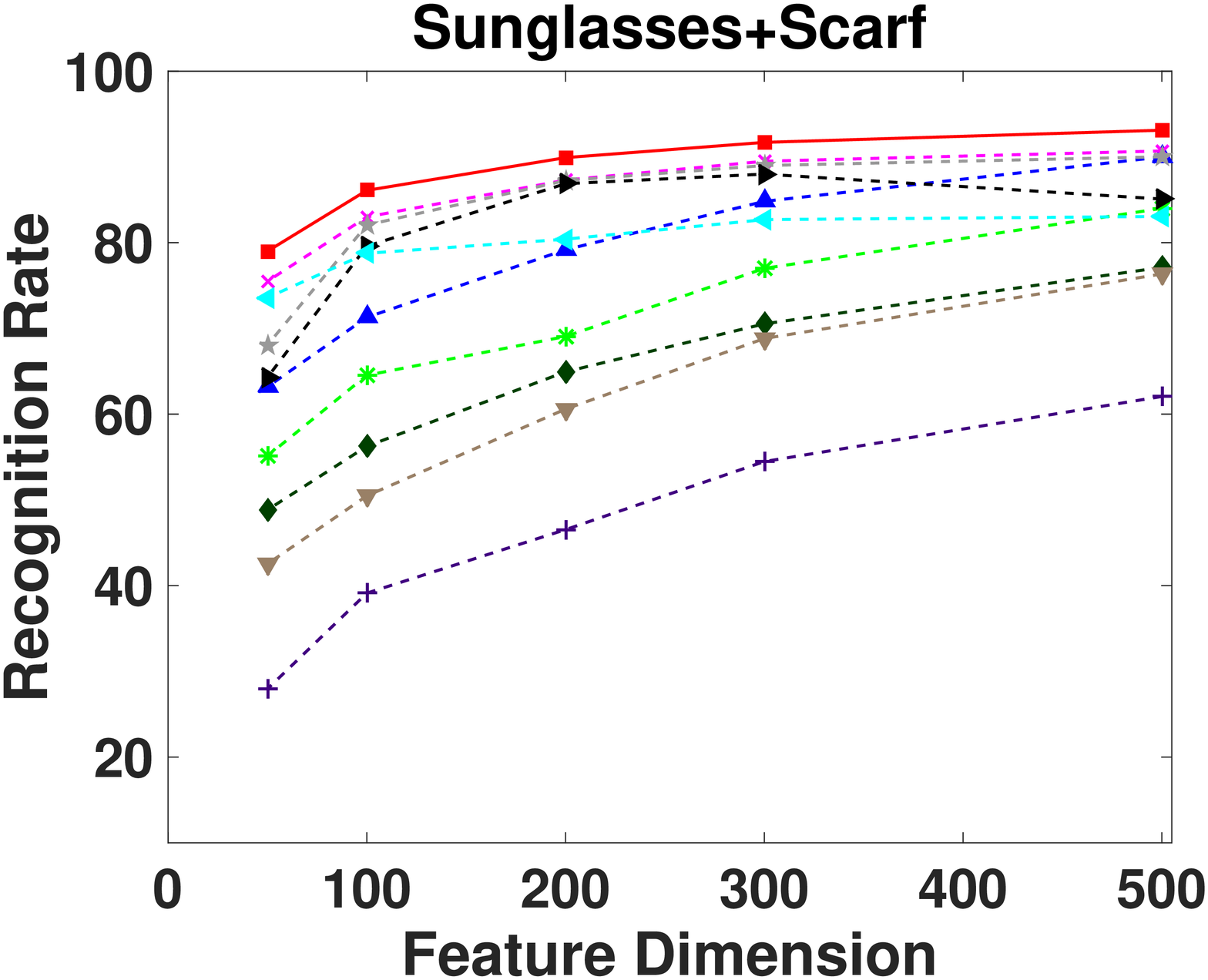}  \label{fig:AR-mixed}}  
\vspace{-1em}
\caption{Accuracy against feature dimensions in three scenarios of AR dataset}
\vspace{-1.5em}
\end{figure} 

In this experiment, we vary feature dimensions from $540$ to $50$ to review the effect of simultaneous dimensionality reduction and occlusion. All compared methods use the raw images as the feature descriptor, except FDDL and D\textsuperscript{2}L\textsuperscript{2}R\textsuperscript{2}, which use random faces \cite{LC-KSVD}, that are generated by projecting a face image onto a random vector. Figs.~\ref{fig:AR-sunglasses}-\ref{fig:AR-mixed} show the recognition rates of JPDL-LR and competing methods under three aforementioned scenarios. Clearly, JPDL-LR achieves higher recognition rate across all dimensions in different scenarios. We can observe that D\textsuperscript{2}L\textsuperscript{2}R\textsuperscript{2} gives good results under occlusion because of LR regularization, when the dimension is still high; however, its performance drops significantly in lower dimensions due to the weakness of random projection. It can be seen that joint DR-DL methods perform better in lower dimensions due to the learned projection matrix. By incorporating local information through supervised neighborhood graph and imposing LR constrain on sub-dictionaries, JPDL-LR is capable of handling noise particularly in low-dimensional data. We also note that the recognition rates of DR methods are remarkably low due to their sensitivity to occlusion. 
\vspace{0.05mm}
 \item \textbf{Extended YaleB Dataset:} This dataset \cite{Yale} contains $2,414$ frontal face images of $38$ human subjects under different illumination conditions. All the face images are cropped and resized to $32 \times 32$ and we randomly select $20$ images per class for training and the rest is used for test. Several images of this dataset can be seen in Fig.~\ref{fig:faces-yale}. In the following experiments, FDDL and D\textsuperscript{2}L\textsuperscript{2}R\textsuperscript{2}, use the Eigenface and all the other methods utilize the raw images as the feature descriptor. 
First, we evaluate the robustness of our method to different levels of pixel and block corruption (from 10\% to 50\%) on YaleB dataset. Fig.~\ref{fig:Yale-pixel} and Fig.~\ref{fig:Yale-block} demonstrate that in noisy scenarios, our method consistently obtains better performance than other methods in all levels of noise and it is mostly followed by D\textsuperscript{2}L\textsuperscript{2}R\textsuperscript{2}. In this experiment, for each level of noise, the projected dimension varies between $1000$ to $50$ and the best achieved result amongst all dimensions is reported. As expected, when the dimension can be high enough, D\textsuperscript{2}L\textsuperscript{2}R\textsuperscript{2} outperforms the joint DR-DL methods on the corrupted dataset. We may infer that, when data are corrupted and dimensionality reduction is not the main goal, existing joint DR-DL methods does not help much in classification; however, JPDL-LR can still obtain better classification results due to LR constraint. It is  interesting that the best performance of our method and D\textsuperscript{2}L\textsuperscript{2}R\textsuperscript{2} is obtained under dimensionality $300$ and $1000$ respectively. These figures also reflect that the performance difference between traditional DR methods and other methods is significant, and this is mainly due to high sensitivity of these methods to illumination changes and corruption.  
Fig.~\ref{fig:Yale-dim-noise} illustrates the recognition rates of all compared methods versus different feature dimensions on 20\% pixel corrupted YaleB dataset. The graph shows that Eigenface, which is used for DR of FDDL and D\textsuperscript{2}L\textsuperscript{2}R\textsuperscript{2}, remarkably fails to preserve the discriminative information in low dimensions; though, the joint DR-DL methods undertake the discrimination of the projected samples in low-dimensions through the joint learning procedure. In particular, the proposed JPDL-LR shows superior performance across all dimensions. 
\end{enumerate}
\subsection{Object Recognition}
In this section, we assess our method on object categorization using COIL-100 \cite{COIL} dataset, which is a testbed for related methods. The COIL dataset contains various views of $100$ objects with different lighting conditions and scales. In our experiments, the images are resized to $32 \times 32$ and the training set is constructed by randomly selecting $10$ images per object from available $72$ images. In addition to alternative viewpoints, we also test the robustness of different methods to noise by adding 10\% pixel corruption to the original images. Some examples of original and corrupted images of COIL dataset can be found in Fig.~\ref{fig:objects-coil}.

\begin{figure}[t]
\centering
\subfloat[]{\includegraphics[width=3.7cm,keepaspectratio]{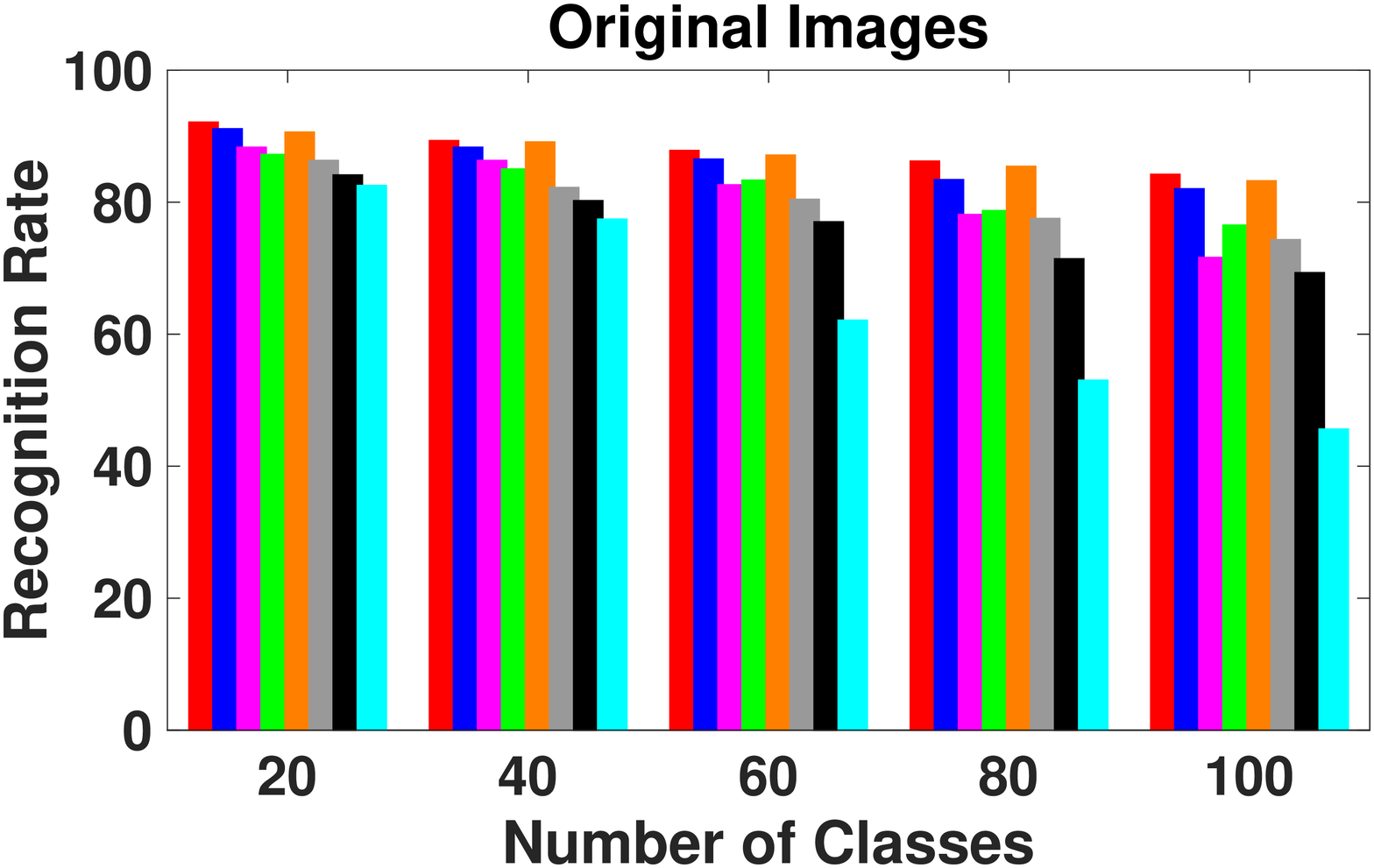} \label{fig:coil-org}}  
\subfloat[]{\includegraphics[width=4.31cm,keepaspectratio]{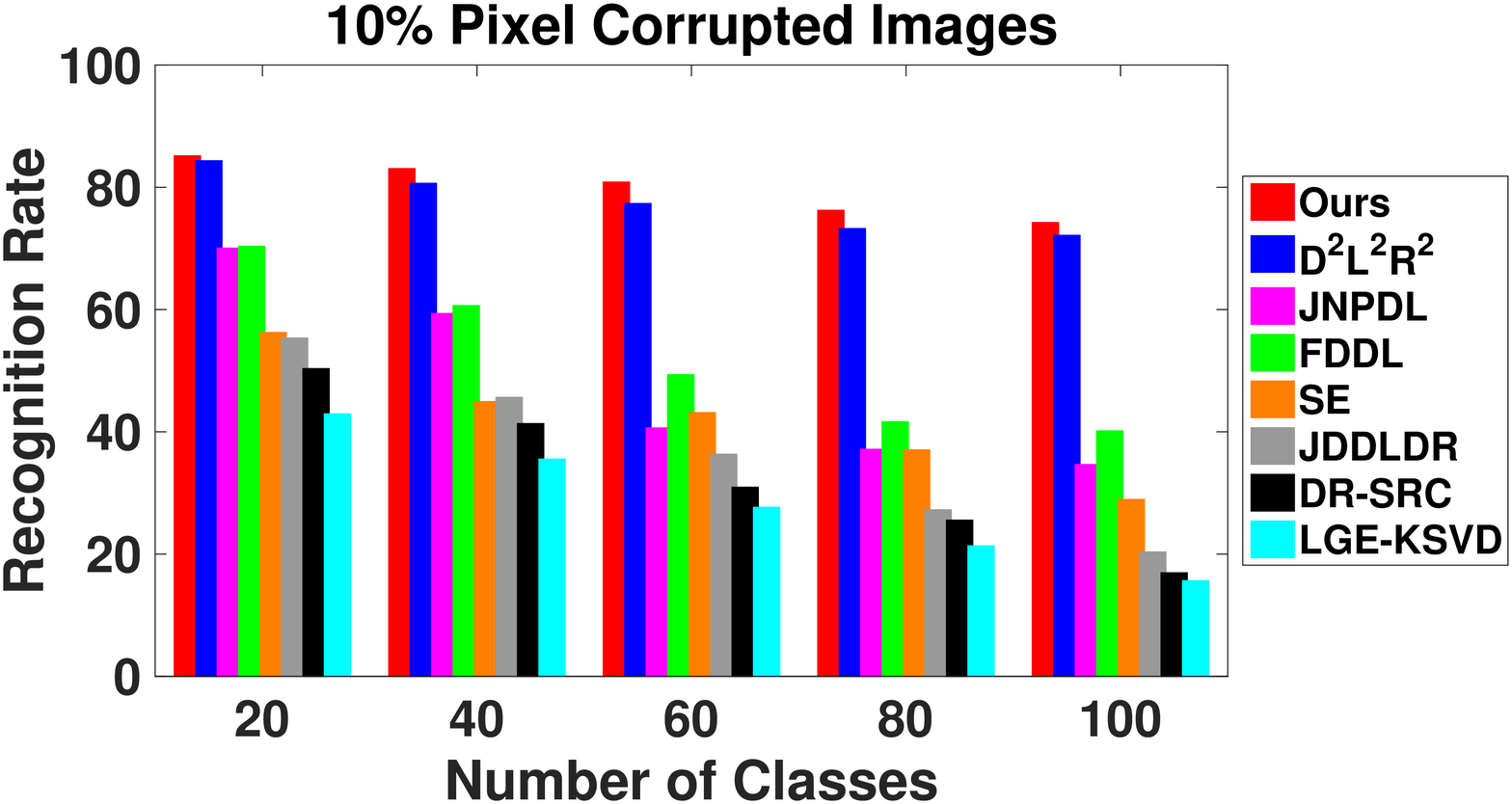} \label{fig:coil-noise10}}  
\subfloat[]{\includegraphics[width=3.5cm,keepaspectratio]{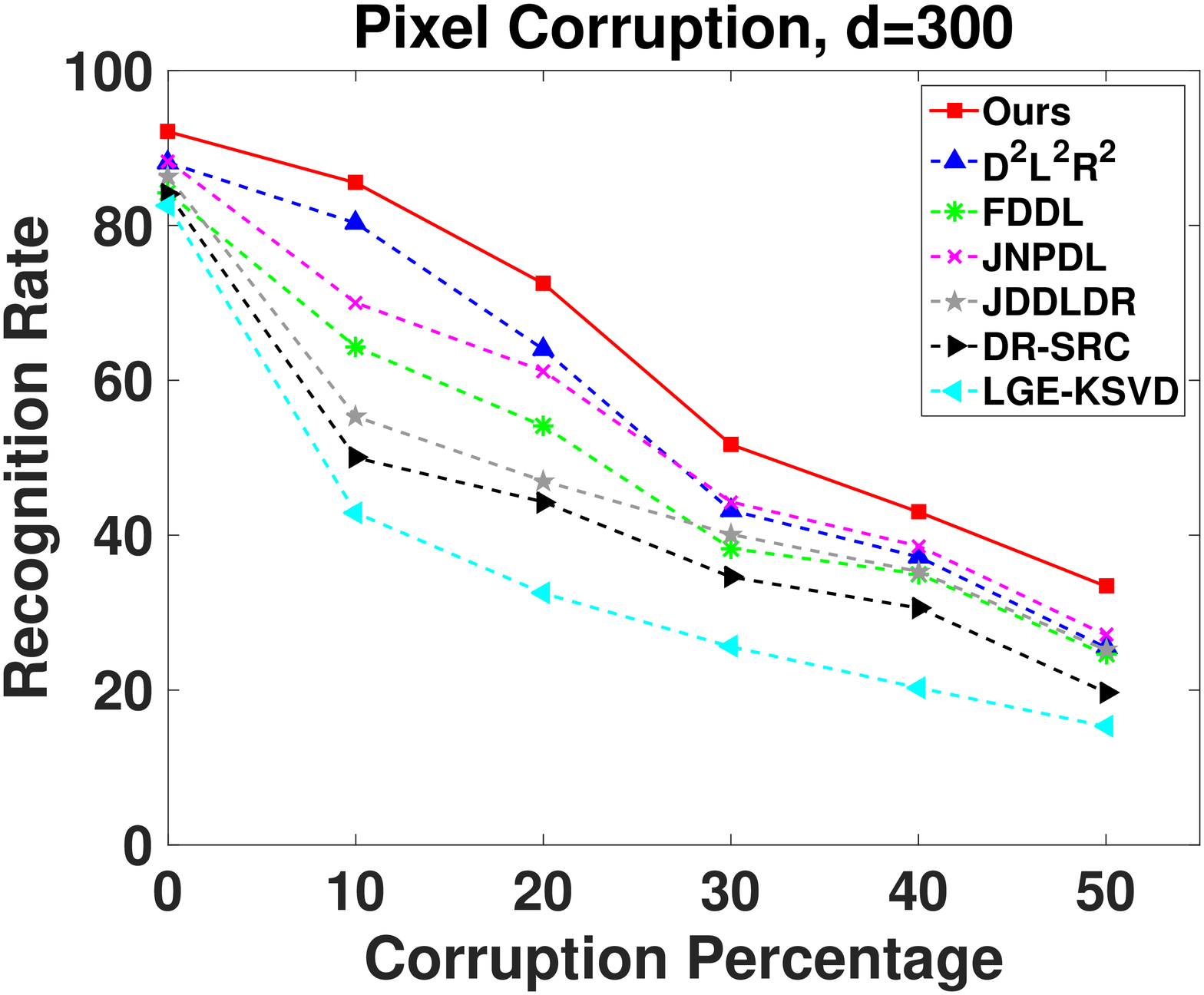}  \label{fig:coil20noise-all-dim300}}  
\vspace{-1em}
\caption{Comparison of accuracy on \textbf{(a)}original images \textbf{(b)}corrupted images of COIL-100 dataset \textbf{(c)}COIL-20 dataset with different levels of noise}
\vspace{-1.5em}
\end{figure}

First, we evaluate the scalability of our method and the competing methods by increasing the number of objects (\ie, classes) from $10$ to $100$. Fig.~\ref{fig:coil-org} and Fig.~\ref{fig:coil-noise10} show the average recognition rates for all compared methods over original images and 10\% pixel corrupted images respectively. Since the traditional DR methods obtained poor results on this dataset, we exclude them from this experiment. Like before, for all the methods, the projected dimension is varied from $1000$ to $50$ and the best achieved performance is reported. It can be observed that the proposed JPDL-LR performs slightly better than the competing methods in the original images; however, when the data are contaminated with noise, this difference becomes more meaningful. When the images are corrupted, all the other methods except D\textsuperscript{2}L\textsuperscript{2}R\textsuperscript{2}, have difficulty obtaining reasonable results. We can see that our method achieves remarkable performance and also demonstrates good scalability. Moreover, we utilize other levels of corruption (from 10\% to 50\%) on COIL-20 dataset and report the results in Fig.~\ref{fig:coil20noise-all-dim300}. In this experiment, we set the feature dimension as $300$; hence, the performance difference between our proposed method and D\textsuperscript{2}L\textsuperscript{2}R\textsuperscript{2} becomes more significant due to our learned projection. Our method totally achieves higher recognition rate than all the competing methods.   

Finally, we design an experiment to show the efficiency of different components of the proposed JPDL-LR framework. To verify the efficacy of LR regularization in the framework, we remove $\alpha \sum_{\substack{i=1}}^K \norm[]{D_i}_{*}$ from Eq.\ref{eq8}. In similar fashion, to evaluate the importance of joint DR-DL, we remove the projection learning part from JPDL-LR, which means that the projection matrix and structured dictionary are learned from training samples separately. We call these two strategies JPDL and JDL-LR respectively and compare them with the proposed JPDL-LR on four datasets in Fig.~\ref{fig:components}. The feature dimension of these methods is set as $0.1$ of the original dimension and the images are corrupted by 40\% pixel noise. We can observe that once the LR regularization is removed, the recognition rate drops significantly in all datasets. Also, we note that JPDL-LR outperforms JDL-LR (with separate projection) and this is mainly due to the fact that some useful information for DL maybe lost in the projection learning phase in this method. The joint learning framework enhances the classification performance, especially when data are highly contaminated and dimension is relatively low.
\section{Conclusion}
\label{sec:conclusion}
In this paper, a novel joint projection and dictionary learning method is proposed. The proposed method simultaneously learns a discriminative projection and dictionary in the low-dimensional space, by incorporating Fisher discrimination criterion, low-rank regularization and supervised graph constraints. These constraints provide the discrimination of projected samples even in highly contaminated environments. When the data contains considerable noise or variation, our method improves the classification performance, especially in lower dimensions.
The experimental results on different benchmark datasets demonstrates the effectiveness of our method for image classification task. Possible future work includes handling larger datasets and extending to non-linear cases.
\par\vfill\par
\clearpage
\bibliographystyle{splncs}
\bibliography{egbib}
\end{document}